\documentclass[sigconf,screen]{acmart}

\usepackage{graphicx} 
\usepackage{algorithm}

\usepackage{amsmath,amssymb}
\usepackage{color}
\usepackage{todonotes} 
\usepackage{multirow}
\usepackage{array}
\usepackage{mathtools}
\usepackage{algpseudocode} 
\usepackage{booktabs}
\usepackage{subfigure}

\usepackage{soul}

\usepackage{colortbl}

\usepackage{lipsum}

\usepackage{mathrsfs}

\usepackage{makecell}
\usepackage{array}

\usepackage{threeparttable}

\usepackage{bm}
\acmSubmissionID{1816}
\definecolor{yellow}{rgb}{0.98,0.82,0.69}
\definecolor{Gray}{gray}{0.95}
\AtBeginDocument{%
  }

\setcopyright{cc}
\copyrightyear{2025}
\acmYear{2025}
\setcctype{by-nc-sa}
\acmConference[MM '25]{Proceedings of the 33rd ACM International Conference on Multimedia}{October 27--31, 2025}{Dublin, Ireland}
\acmBooktitle{Proceedings of the 33rd ACM International Conference on Multimedia (MM '25), October 27--31, 2025, Dublin, Ireland}
\acmDOI{10.1145/3746027.3755038} 
\acmISBN{979-8-4007-2035-2/2025/10}

\begin{document}

\title{ReactDiff: Fundamental Multiple Appropriate Facial Reaction Diffusion Model}


\author{Cheng Luo}
\affiliation{%
  \institution{Monash University}
\city{Melbourne}
  \country{Australia}
}
\email{cheng.luo@monash.edu}

\author{Siyang Song}
\authornote{Corresponding Author}
\affiliation{%
\institution{University of Exeter}
\city{Exeter}
  \country{United Kingdom}
}
\email{s.song@exeter.ac.uk}

\author{Siyuan Yan}
\affiliation{%
  \institution{Monash University}
  \city{Melbourne}
  \country{Australia}
}
\email{siyuan.yan@monash.edu}

\author{Zhen Yu}
\affiliation{%
  \institution{Monash University}
  \city{Melbourne}
  \country{Australia}
}
\email{zhen.yu@monash.edu}

\author{Zongyuan Ge}
\affiliation{%
  \institution{Monash University}
  \city{Melbourne}
  \country{Australia}
}
\email{zong.yuan@monash.edu}

\renewcommand{\shortauthors}{Cheng Luo, Siyang Song, Siyuan Yan, Zheng Yu, and Zongyuan Ge}
\begin{teaserfigure}
\centering
\begin{minipage}[t]{\linewidth}
\centering
{\fontsize{10pt}{10pt}\selectfont \textbf{Project Page:} \url{https://reactdiff.github.io}}\\
\end{minipage}   
\end{teaserfigure}
\begin{abstract}
The automatic generation of diverse and human-like facial reactions in dyadic dialogue remains a critical challenge for human-computer interaction systems. Existing methods fail to model the stochasticity and dynamics inherent in real human reactions. To address this, we propose ReactDiff, a novel temporal diffusion framework for generating diverse facial reactions that are appropriate for responding to any given dialogue context. Our key insight is that plausible human reactions demonstrate smoothness, and coherence over time, and conform to constraints imposed by human facial anatomy. To achieve this, ReactDiff incorporates two vital priors (spatio-temporal facial kinematics) into the diffusion process: i) temporal facial behavioral kinematics and ii) facial action unit dependencies. These two constraints guide the model toward realistic human reaction manifolds, avoiding visually unrealistic jitters, unstable transitions, unnatural expressions, and other artifacts. Extensive experiments on the REACT2024 dataset demonstrate that our approach not only achieves state-of-the-art reaction quality but also excels in diversity and reaction appropriateness. Our code is publicly available at \url{https://github.com/lingjivoo/ReactDiff}.
\end{abstract}

\ccsdesc[500]{Human-centered computing~User interface programming}
\ccsdesc[500]{Computing methodologies~Computer vision tasks}

\keywords{Human-computer Interaction, Human Behavior Understanding, Video Generation}

\maketitle

\section{Introduction}

\begin{figure}[t!]
    \centering
    \includegraphics[width=0.88\columnwidth]{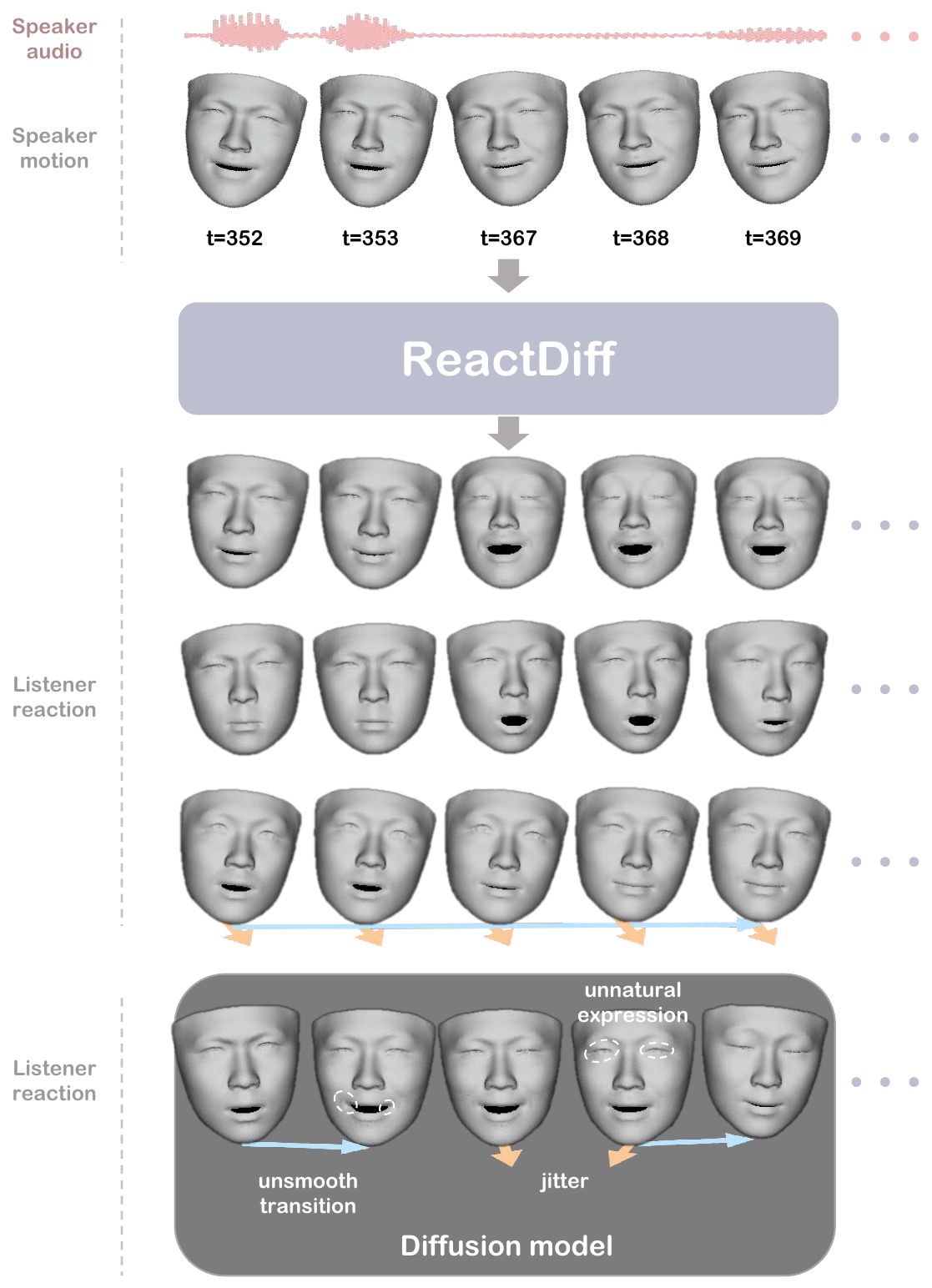} 
        \caption{\textbf{Demonstration} of diverse reactions generated by ReactDiff and \textbf{Limitations} of standard diffusion model for \emph{online} facial reaction prediction.}
    \label{fig:intro}
\end{figure}

\noindent A long-standing goal in artificial intelligence (AI) is enabling intelligent agents to precisely comprehend intentions and emotions conveyed via human expressive audiovisual behaviors, and in turn, respond to human-like verbal and non-verbal behaviors during human-computer interactions \cite{brave2007emotion}. Although large language models (LLMs) \cite{thoppilan2022lamda,  brown2020language} have fueled groundbreaking advancements in language-based verbal communication interfaces, automatic agents capable of expressing realistic and contextual appropriate human-style facial behaviors (reactions) in response to different user behaviors still remain underexplored.

While early deterministic facial reaction generation models  \cite{huang2018generating,huang2018generative,song2022learning,shao2021personality,nojavanasghari2018interactive,woo2021creating,woo2023amii} attempted to reproduce the real facial reaction (called GT facial reaction) specifically expressed by the individual for responding to the input behavior (called speaker behavior), their training typically suffer from `one-to-many' problem as individuals may react differently to the same speaker behavior due to varied factors (\emph{e.g.}, personality \cite{song2022learning}), i.e., these facial reactions all remain contextually appropriate in response to the speaker behavior \cite{mehrabian1974approach,song2023multiple}. The recently emerged online multiple appropriate facial reaction generation (MAFRG) task \cite{song2023multiple} aims to \textit{generate multiple diverse facial reactions that individuals would naturally and appropriately display in response to any given speaker behavior in real-time}. This task is challenging, as appropriate facial reactions (AFRs) should be adaptive to the given speaker behavior at various levels, spanning from the speaker's voice, tone, expressions, and appearance \cite{song2022learning}, to unanticipated behavioral changes and contexts \cite{mehrabian1974approach} in the interaction. As a result, recent solutions frequently represent multiple AFRs triggered by each speaker behavior as a Gaussian-style distribution in a continuous \cite{luo2024reactface,xu2023reversible} or discrete \cite{ng2022learning,liang2023unifarn} latent space, preventing their training from ill-posed `one-to-many mapping' problem. 

This way, multiple different AFRs can be sampled by the obtained distribution. However, since the spontaneous AFRs for responding to different speaker behaviours in real-world scenarios can show varied and complex distribution, such learned Gaussian AFR latent distributions may struggle to effectively represent them.

Alternatively, the denoising diffusion model (DDM) can effectively model various real data distribution through denoising processes \cite{ho2020denoising, song2020score}, and thus can well address limited diversity issues. 
As a result, some recent studies \cite{zhu2024perfrdiff,nguyen2024vector,yu2023leveraging} have specifically explored diffusion-based MAFRG solutions, which directly apply standard diffusion strategy to generate AFRs from reference images.
These diffusion-based offline or online MAFRG models \cite{zhu2024perfrdiff,yu2023leveraging} continuously generate short AFR segments conditioned on the current and previously expressed speaker behaviors to form the entire facial reaction sequence. However, the AFRs generated by such standard DDM-based online MAFRG models suffer from noticeable jitters, incoherent transitions between facial reaction segments, and unnatural expressions (shown in Fig.~\ref{fig:intro}). \textit{\textbf{This is because the standard DDM does not consider crucial priors of human facial behavioral kinematics nor specifically account for previously generated AFRs and speaker behaviors within the diffusion process.}}

To well adapt the powerful DDM to the online MAFRG task, this paper proposes the first online real-time MAFRG diffusion strategy called ReactDiff, which addresses the above fundamental issues by restructuring the architecture of standard DDM. 
Specifically, 
our ReactDiff incorporates temporal cues (with the global timestamp of the conversation and historical information) to obtain facial reactions with reasonable (not disordered) and consecutive changes over time. 
A facial behavioral kinematics constraint is then proposed to regulate the pace of expression and pose changes, aligning them with natural human behavioral rhythms that avoid extremes of being too slow or rapid. To obtain natural facial expressions and movements that adhere to human facial anatomy, we summarize relationships between individual facial muscle movements (facial action units) and enforce expert rules to correct unusual facial movements in the generated reactions. These modifications introduce crucial inductive biases into the model, steering the diffusion model toward realistic human facial behavior dynamics.
Our main contributions are summarized as follows:
\begin{itemize}
\item We propose a temporal reaction diffusion model to generate diverse and naturalistic reactions online in response to speaker behaviors.
\item We introduce two novel constraints that enable diffusion models to learn distributions of reactions aligned with human facial behavioral kinematics and facial expressions.
\item Extensive experiments showcase that our model largely outperforms state-of-the-art methods in terms of diversity, appropriateness, and realism of the generated facial reactions.
\end{itemize}

\section{Related Work}
\label{sec:related_work}

\textbf{Automatic Facial Reaction Generation.} Facial reaction generation aims to predict human facial reactions (including expressions and head poses) in response to the currently given non-verbal and verbal signals conveyed by the conversational partner (speaker).
Many prior approaches have been developed with the primary aim of replicating the ground truth ('GT') facial reactions by the corresponding listener in specific contexts. For instance, Huang et al. \cite{huang2017dyadgan,huang2018generating} utilized a conditional Generative Adversarial Network \cite{mirza2014conditional,goodfellow2020generative} to generate the listener's authentic facial reaction sketch based on the speaker's facial action units (AUs). Similar frameworks \cite{huang2018generating,huang2018generative,song2022learning,shao2021personality,nojavanasghari2018interactive,woo2021creating,woo2023amii} extended these methods by incorporating additional modalities (\emph{e.g.}, audio and textual features) as inputs. However, these deterministic approaches 
often converge to generate average facial reactions \cite{song2023multiple}.
Ng et al. \cite{ng2022learning} proposed a non-deterministic method capable of generating different facial reactions to the same speaker behavior, yet still remained producing reactions with similar patterns.
To tackle this issue, recent studies \cite{xu2023reversible,luo2024reactface,liang2023unifarn} re-framed the 'one-to-one mapping' training strategy into an 'one-to-many' supervision. However, their architectures limit the complex distribution modeling.
As an effective tool to model any data distribution, diffusion models have superior ability to sample appropriate reactions, and their sampling solvers consider independent stochasticity.

\textbf{Diffusion Models.} Denoising diffusion or score-based generative models \cite{song2020score, ho2020denoising} have emerged as powerful deep learning frameworks for various data synthesis tasks (\emph{e.g.,} image \cite{dhariwal2021diffusion,rombach2022high}, 3D shape \cite{luo2021diffusion} and human motion \cite{tevet2022human,zhang2022motiondiffuse,barquero2022belfusion} synthesis).
These frameworks progressively diffuse each real data point with random noise (called diffusion process), which can be mathematically described by either a stochastic differential equation (SDE) or an ordinary differenial equation (ODE) \cite{song2020score}. Then, a network is learned to reverse this diffusion process by removing noise corruptions added to the data. Specifically, the SDE solver-based reverse diffusion considers more stochastic factors in generation compared to the deterministic sampling via an ODE solver. Subsequent investigations \cite{peebles2023scalable, rombach2022high} on the applications of diffusion models have unveiled their strengths in scalability and seamless integration with diverse forms of conditions such as text \cite{kawar2023imagic,rombach2022high}, pose \cite{zhang2023adding}, action \cite{tevet2022human}, dense maps \cite{zhang2023adding,ji2023ddp} and semantics maps \cite{zhang2023adding,ji2023ddp}. In comparison to conditional Generative Adversarial Networks \cite{mirza2014conditional} and Variational Autoencoders \cite{kingma2013auto, van2017neural}, diffusion models with classifier-free guidance technique \cite{ho2022classifier} show greater potential in incorporating multi-modal conditions while inducing less harm to the generation process.

\begin{figure*}[t!]
    \centering
    \includegraphics[width=2\columnwidth]{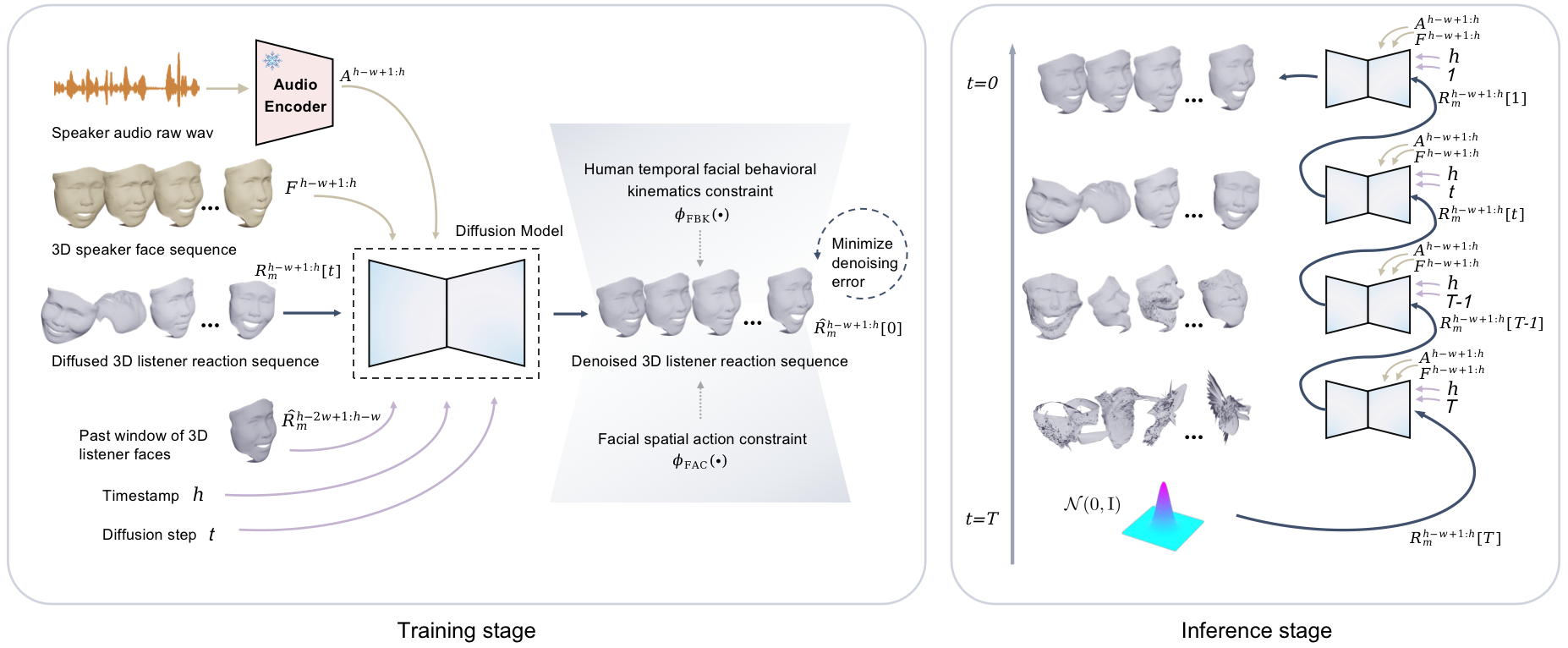} 
        \caption{\textbf{Overview} of the proposed ReactDiff model. \textbf{Left:} the training stage of ReactDiff, wherein ReactDiff is learned to denoise 3D listener reaction sequence with given conditions and two constraints. \textbf{Right:} the inference stage of ReactDiff, involving the sampling of reaction sequences through multiple reverse diffusion steps.}
    \label{fig:overview}
\end{figure*}

\section{Preliminary and Problems}

\noindent Diffusion models are latent variable models that model the real data $x[0]$ as Markov chains $\{x[T], \cdots, x[0] \}$.
Specifically, the forward diffusion process of standard diffusion models is achieved by incrementally injecting a series of Gaussian noise to the input clean data $x[0]$ to encode it as $q(x[0])$, which can be formulated as:
\begin{equation}
    q(x[1:T] | x[0]) = \prod^T_{t=1} q(x[t] | x[t-1])
\end{equation}
where $x[1:T]$ represents $T$ noisy data samples obtained from the denoising step $t= 0$ to $t = T$. Subsequently, a reverse denoising process is achieved by a denoiser network $p_\theta$ that incrementally denoises the diffused samples $x[T:1]$ to recover the original clean data $x[0]$ as:
\begin{equation}
p_{\theta}(x[0:T]) = p(x[T])\prod^T_{t=1} p_{\theta}(x[t-1]|x[t])
\end{equation}

\noindent In the \emph{offline} MAFRG setting, the diffusion model generates the entire sequence of each AFR $\mathcal{R}^{1:H}_m[0]$ at once, covering the full time span $1$ to $H$. In contrast, the \emph{online} MAFRG task requires to iteratively produce either a single AFR frame $\mathcal{R}^{h}_m[0]$ for each $h \in [1:H]$, or a short AFR segment $\mathcal{R}^{h-w+1 : h}_m[0]$, where $w$ denotes the window length. This streaming nature imposes extra challenges: ensuring consistency between facial behaviours expressed in consecutive time windows, as any discontinuity would be highly noticeable. Moreover, directly diffusing AFRs in each window from a random noise can introduce semantic inconsistencies and abrupt transitions in facial frames across window boundaries.

\noindent Existing \emph{online} MAFRG diffusion models (\emph{e.g.}, \cite{zhu2024perfrdiff,yu2023leveraging}) generate each AFR segment $\mathcal{R}^{h-w+1 : h}_m[0]$ solely from the corresponding speaker facial and audio behaviors $F^{h-w+1:h}$ and $A^{h-w+1:h}$ observed in the same time window $[h-w+1, h]$ as:
\begin{equation}
    p_\theta(\mathcal{R}^{h-w+1:h}_m[t-1] | \mathcal{R}^{h-w+1:h}_m[t], F^{h-w+1:h}, A^{h-w+1:h}) 
\end{equation}
where ${p}_{\theta}(\cdot)$ denotes their diffusion model denoising current AFR segments $\mathcal{R}^{h-w+1:h}_m$ conditioned on $F^{h-w+1: h}_m$ and $A^{h-w+1: h}_m$. The key limitation of these methods is that their failure to account for crucial temporal facial behavioral kinematics or spatial relationships of facial muscle movements within the diffusion denoising process (these are instead handled by separate components such as LSTM \cite{nguyen2024vector} or subsequent linear layers \cite{yu2023leveraging}), leading them fail to maintain the temporal coherence between previously and currently generated facial reactions nor generate plausible facial displays.

\section{Methodology}
\label{method}

\noindent This section presents our ReactDiff model for the online MAFRG task, which integrates the natural \textbf{human temporal facial behavioral kinematics}~$\phi_{\text{FBK}}(\hat{\mathcal{R}}^{h-2w+1, h}_m)$ and \textbf{spatial facial action dependencies}~$\phi_{\text{FAC}}(\hat{\mathcal{R}}^{h-w+1, h}_m)$ into the standard diffusion process to form a human facial behavior-specific diffusion strategy, where $\phi_{\text{FBK}}(\hat{\mathcal{R}}^{h-2w+1, h}_m)$ ensures that each current appropriate facial reaction (AFR)~$\hat{\mathcal{R}}^{h-w+1, h}_m$ remains temporal continuous with the corresponding previously generated AFR segment $\hat{\mathcal{R}}^{h-2w+1, h-w}_m$, while $\phi_{\text{FAC}}(\hat{\mathcal{R}}^{h-w+1, h}_m)$ prevents generating unrealistic facial displays. Through these constraints, our ReactDiff generates multiple ($M$) distinct human-like AFR segments expressed for current temporal window $[h-w+1, h]$ (see \cite{song2023multiple} for the definition) as:
\begin{equation}
    \mathcal{\hat{R}}^{h-w+1: h} = \bigl\{\hat{\mathcal{R}}^{h-w+1: h}_1,\;\hat{\mathcal{R}}^{h-w+1: h}_2,\;\cdots,\;\hat{\mathcal{R}}^{h-w+1: h}_M \bigr\},
\end{equation}
where each segment $\hat{\mathcal{R}}^{h-w+1: h}_m = \{\hat{r}^\tau_m\}_{\tau=h-w+1}^{h}$ represents a short face video sequence comprising $w$ frames. Here, at the timestamp $\tau \in [h-w+1, h]$, the AFR frame $\hat{\mathcal{R}}^\tau_m \in \hat{\mathcal{R}}^{h-w+1: h}_m$ is dynamically and adaptively generated to respond to the current multi-modal speaker behavior characterized by $w$ facial behavior frames $F^{h-w+1:h} = \{f^\tau\}_{\tau=h-w+1}^h$ and the corresponding auditory signal $A^{h-w+1:h} = \{a^\tau\}_{\tau=h-w+1}^h$. Our diffusion-based denoising process can be formally summarized as:
\begin{equation}
\begin{aligned}
p_\theta\bigl(
\hat{\mathcal{R}}^{h-w+1:h}_m[t-1] &\;\big\vert\;
\hat{\mathcal{R}}^{h-w+1:h}_m[t],\;
F^{h-w+1:h}, \; 
A^{h-w+1:h}, \\
&\phi_{\text{FAC}}\bigl(
\hat{\mathcal{R}}^{h-w+1:h}_m[t]
\bigr), 
\phi_{\text{FBK}}\bigl(\hat{\mathcal{R}}^{h-2w+1:h}_m[t]\bigr)
\bigr),
\end{aligned}
\end{equation}
where 
$\hat{\mathcal{R}}^{h-2w+1:h}_m$ represents the predicted AFR frames from the preceding temporal window to the current window, \emph{i.e.}, $\phi_{\text{FBK}}$ constraints the denoising process for generating current AFR segments based on previously generated AFR segment, ensuring the temporal coherence between them, while $\phi_{\text{FAC}(\cdot)}$ constraints spatial facial action dependencies to ensure the realism of each generated facial display. In this paper, each input speaker facial behavior frame $f^\tau \in F^{h-w+1:h}$ is represented by a set of 3DMM coefficients capturing both facial expression and head pose. Following \cite{shao2021personality,song2022learning,ng2022learning}, we use a small time window $w$, reflecting the time delay introduced by human cognitive processes~\cite{card1986model}. An overview of the entire ReactDiff pipeline is shown in Fig.~\ref{fig:overview}.

\subsection{Spatio-temporal Dependency-aware Online Facial Reaction Diffusion}

\noindent Since online MAFRG requires to continuously generate short AFR frames/segments to form each whole AFR video, our ReactDiff generates multiple but different AFR segments $\hat{\mathcal{R}}^{h-w+1: h} = \{\hat{\mathcal{R}}^{h-w+1: h}_1,  \\ 
\hat{\mathcal{R}}^{h-w+1: h}_2, \cdots, \hat{\mathcal{R}}^{h-w+1: h}_M\}$ in current interval $[h-w+1, h]$, where each $\hat{\mathcal{R}}^{h-w+1: h}_m = \{ r^\tau_m \}^{h}_{\tau=h-w+1}$ consisting of $w$ frames is produced based on not only the current speaker audio-visual behaviors $F^{h-w+1:h}$ and $A^{h-w+1:h}$ but also facial spatial dependency $\phi_{\text{FAC}}(\hat{\mathcal{R}}^{h-w+1:h}_m)$ and temporal dependency $\phi_{\text{FBK}}(\hat{\mathcal{R}}^{h-2w+1:h}_m)$ considering previous facial reactions. This can be formulated as learning the joint probabilistic distribution for generating AFR segments at the time interval $[h-w+1:h]$ as: 
\begin{equation}
\begin{aligned}
  p(  & \mathcal{R}^{h-w+1:h}_m | \mathcal{R}^{h-2w+1:h-w}_m,  \phi_{\text{FAC}}(\mathcal{R}^{h-w+1:h}_m), \phi_{\text{FBK}}(\mathcal{R}^{h-2w+1:h}_m),  \\  &  F^{h-w+1:h}, A^{h-w+1:h}, t, h) \\ = & p(\mathcal{R}_m^{h-w+1:h}[T]) \prod_{t=1}^{T}p(\mathcal{R}^{h-w+1:h}_{m}[t-1] |\mathcal{R}^{h-w+1:h}_{m}[t],  \\  & \mathcal{R}^{h-2w+1:h-w}_{m}[0], \phi_{\text{FAC}}(\mathcal{R}^{h-w+1:h}_m [t]), \phi_{\text{FBK}}(\mathcal{R}^{h-2w+1:h}_m[t] ),
  \\ &  F^{h-w+1:h}, A^{h-w+1:h}, t, h)
\end{aligned}
\label{eq:dis}
\end{equation}
where $t$ denotes the diffusion step index; $T$ represents the number of total diffusion steps; $h$ is the temporal timestamp, while $\phi_{\text{FAC}}(\mathcal{R}^{h-w+1:h}_m[t])$ and $\phi_{\text{FBK}}(\mathcal{R}^{h-2w+1:h}_m[t])$ acting as joint spatio-temporal constraints during the current AFR segment distribution learning.

\textbf{Injecting spatio-temporal constraints into diffusion:} While diffusion models demonstrate substantial potential in modeling the distribution of AFRs conditioned on given speaker behaviors, they cannot explicitly understand the underlying human facial temporal kinematics and spatial action constraints when synthesizing facial behaviors. Facial reaction diffusion models without encoding expression priors tend to mimic the average mode of training facial reactions. This mean distribution may cover abnormal facial behaviors. As a result, facial reactions generated from general diffusion models may suffer from issues such as jitters, unstable transitions between frames, and unnatural human facial behaviors, making them implausible and unrealistic.
To enforce our diffusion model to generate human-like and realistic AFRs, we inject spatial and temporal constraints into our ReactDiff's forward propagation process via our classifier-free~\cite{ho2022classifier} training strategy. During training (left portion of Fig.~\ref{fig:overview}), we gradually inject Gaussian noise into each real AFR segment $\mathcal{R}^{h-w+1:h}_m[0]$ (real AFR expressed by human listener) that responds to the given speaker behavior, resulting in a diffused real AFR segment $\mathcal{R}^{h-w+1:h}_m[t]$. This forward diffusion process can be formulated as: $q_t(\mathcal{R}^{h-w+1:h}_m[t]|\mathcal{R}^{h-w+1:h}_m[0])$.

Subsequently, a network is employed to eliminate the added noise, yielding a denoised AFR segment $\mathcal{R}^{h-w+1:h}_m[0]$ conditioned on auditory signal $A^{h-w+1:h}$ and facial behavior $F^{h-w+1:h}$ expressed by the speaker, temporal timestamp $h$, as well as the previously predicted AFR segment $\mathcal{R}^{h-2w+1:h-w}_m[0]$. In this training process, the denoised AFR segment and changes of predicted noise could be used for spatial and temporal constraint, making the denoising model learn the distribution towards natural human facial reactions with coherent variations over time.

While MAFRG involves generating multiple AFRs in response to each speaker behavior, we further employ classifier-free guidance~\cite{ho2022classifier}. Instead of directly predicting each AFR, our ReactDiff estimates a score function $\nabla_{\mathcal{R}^{h-w+1:h}_{m}[t]} \text{log}q_t(\mathcal{R}^{h-w+1:h}_{m}[t])$ through a learned network structured as a U-Net architecture \cite{ronneberger2015u}. With the estimated score function, ReactDiff can sample AFRs through reverse-time SDE, which incorporates stochasticity in the denoising process (more details are provided in Appendix~\ref{sec:ode_sde}). This way, ReactDiff is meticulously optimized to match the score with the objective as:
\begin{equation}
\begin{aligned}
 \mathcal{L}_\text{dm} = &
    \mathbb{E}_{\mathcal{R}^{h-w+1:h}_{m}[0], t, \epsilon \sim \mathcal{N}(\mathrm{0}, \mathrm{I})} \| 
    p_{\theta}(\mathcal{R}^{h-w+1:h}_{m}[0] + \sigma_t  \epsilon, F^{h-w+1:h},
    \\ &  
    A^{h-w+1:h}, t, h) -  \nabla_{\mathcal{R}^{h-w+1:h}_{m}[t]} \text{log}q_t(\mathcal{R}^{h-w+1:h}_{m}[t]) \|^2_2.
\end{aligned}
\end{equation} 
where $\epsilon$ denotes noise from the Gaussian distribution $\mathcal{N}(\mathrm{0}, \mathrm{I})$. This objective optimizes the denoising network to predict the noise $p_{\theta}(\mathcal{R}^{h-w+1:h}_{m}[0] + \sigma_t  \epsilon, F^{h-w+1:h}, A^{h-w+1:h}, t, h)$ to be close to the injected one $\nabla_{\mathcal{R}^{h-w+1:h}_{m}[t]} \text{log}q_t(\mathcal{R}^{h-w+1:h}_{m}[t])$ . Once we have learned the score-matching network $p_\theta$, we can derive an empirical estimation of SDE and solve it via a numerical solver. Through this reverse-time diffusion process by the SDE solver, we obtain the a solution trajectory $ \{\hat{\mathcal{R}}^{h-w+1:h}_{m}[t]\}_{t=0}^T$ from denoising step $t= T$ to $t= 0$, as depicted in the right part of Fig.~\ref{fig:overview}. Consequently, $\hat{\mathcal{R}}^{h-w+1:h}_{m}[0]$ can be regarded as an approximate sample drawn from the AFR distribution $q_0(\mathcal{R}^{h-w+1:h}_{m}[0])$ in response to current speaker behavior.

\subsection{Spatio-temporal Facial Kinematics}
\label{sec:facial_constraint}

\noindent We formulate our spatial and temporal facial constraints in the form of two critical loss terms: the human temporal facial behavioral kinematics constraint loss and the spatial facial action constraint loss to ensure our ReactDiff model being aware of such constraints during its facial reaction diffusion process.

\noindent \textbf{Human temporal facial behavioral kinematics constraint $\phi_{\text{FBK}(\cdot)}$:} The  human temporal facial behavioral kinematics constraint loss $\mathcal{L}_\text{fbk}$ is introduced to enforce our ReactDiff generating temporally coherent AFRs, \emph{i.e.}, regulating facial behavior changes over time to ensure they are plausible to be expressed by human beings, This is achieved by the joint optimization of the score matching network (denoising network) as:
\begin{equation}
\begin{aligned}
 \mathcal{L}_\text{fbk} = \sum_{i=h-w+2}^{h} & \|v^{i \leftarrow i-1}_{m}[t] - \hat{v}^{i \leftarrow i-1}_{m}[t]  \| \\ + & \|v^{i \leftarrow i-w}_{m}[t] - \hat{v}^{i \leftarrow i-w}_{m}[t]  \|
\end{aligned}
\end{equation}
where $v^{i \leftarrow i-1}_{m}[t] =  \|\nabla_{r^{i}_{m}[t]} \text{log}q_t(r^{i}_{m}[t]) - \nabla_{r^{i-1}_{m}[t]} \text{log}q_t(r^{i-1}_{m}[t]) \|$ is represented as the velocity score function at the time $i$, while $\hat{v}^{i \leftarrow i-1}_{m}[t] = \|p_\theta(r_{m}^i[t], c) - p_\theta(r_m^{i-1}[t], c) \|$ denoting the change velocity between scores estimated for two adjacent generated AFR frames. For ease of the presentation, we represent all conditions (\emph{e.g.}, $F^{h-w+1:h}, A^{h-w+1:h}, t, h$ and past frames) as $c$ in the following contents. In particular, $v^{i \leftarrow i-w}_{m}[t] = \| \nabla_{r^{i}_{m}[t]} \text{log}q_t(r^{i}_{m}[t]) - \nabla_{r^{i-w}_{m}[t]} \text{log}q_t(r^{i-w}_{m}[t]) \| / w$ denotes the velocity score between two temporally neighboring real AFR segments, while $\hat{v}^{i \leftarrow i-w}_{m}[t] = \|p_\theta(r_m^i[t], c) - p_\theta(r_m^{i-w}[t], c) \| / w$ expressing the estimated velocity change score between two temporally neighboring generated AFR segments. This constrains the differences between temporally adjacent generated AFR segments to be coherent as temporally adjacent real AFR segments. By constraining the diffusion model based on these velocity terms, the model enforces the temporal patterns of the generated AFRs to align with the velocity of changes (temporal patterns) of real human facial behaviors. Here, we found that facial reactions synthesized in early diffusion steps, where diffusion noise levels are high, exhibit minimal facial movements. Consequently, enforcing \emph{facial kinematics constraints} too early in the denoising process could inadvertently push reactions away from the true data distribution. To deal with this issue, we follow a scheduling strategy \cite{yuan2023physdiff} that introduces the constraint in the later steps (from $t=5$ to $t=0$) of the denoising process.

\begin{figure}[t]
    \centering
    \includegraphics[width=0.9\columnwidth]{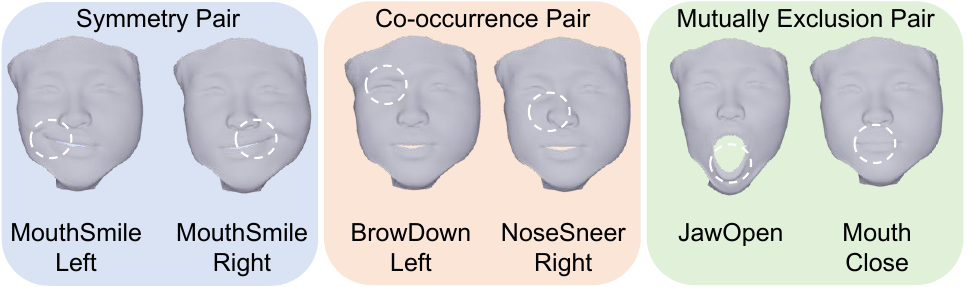} 
\caption{\textbf{Illustration} of three types of facial AU relationships.}
\label{fig:fac_pairs}
\end{figure}

\noindent \textbf{Facial spatial action constraint $\phi_{\text{FAC}}$:} 
While the kinematic constraint can not prevent our ReactDiff from generating unnatural spatial facial expressions (\emph{i.e.}, expressions seldom observed in real human-human interactions), we propose a facial spatial action loss to constraint spatial relationships among facial muscle activations. 
Specifically, we introduce three types of dependencies between facial actions according to previous facial action unit detection studies \cite{luo2022learning,zhang2018weakly,li2019semantic} and a facial psychology study \cite{ekman1978facial}, including 
\textbf{symmetric}, \textbf{co-occurred}, and \textbf{mutually exclusive} AU pairs. For instance, considering facial topology, 'MouthSmileLeft' and 'MouthSmileRight' are recognized as a pair of symmetry action units. 
Similarly, pairs such as 'BrowDownLeft' and 'BrowDownRight', and 'CheekSquintLeft' and 'CheekSquintRight', all present symmetrical behaviors. Furthermore, we identify pairs of action units with high co-occurrence probabilities, such as 'NoseSneerRight' and 'BrowDownLeft', and 'MouthDimpleLeft' and 'MouthClose'. Besides, we conclude pairs of facial actions displaying mutually exclusive behaviors, including 'MouthSmileLeft', 'MouthFrownLeft', 'JawOpen' and 'MouthClose'. To characterize such spatial facial action relationships, we compute the differences between each facial action unit (AU) pair (\emph{i.e.}, facial expression coefficients), which constraints AU pairs in the generated AFR frame to match the spatial patterns in observed real human facial expressions. This can be formulated as:
\begin{equation}
\begin{aligned}
 \mathcal{L}_\text{fac} & =   \sum_i \sum_{j=i+1} \underbrace{ \mathbf{1}_{\Omega_\text{sym}}(i,j) \|  d_{i,j}  -  \hat{d}_{i,j} \|}_{\text{symmetry}}  \\  & + \underbrace{\mathbf{1}_{\Omega_\text{coo}}(i,j) \|  d_{i,j}  -  \hat{d}_{i,j} \|  }_{\text{co-occurrence}}   +
 \underbrace{ \mathbf{1}_{\Omega_\text{exc}}(i,j) \|d_{i,j}  -  \hat{d}_{i,j} \| }_{\text{mutually exclusion}} 
\end{aligned}
\end{equation}
where $\Omega_{\text{sym}}$, $\Omega_{\text{coo}}$ and $\Omega_{\text{exc}}$ represent indicator functions describing three sets of AU pairs defining AU pairs whose relationships are symmetric, co-occurred and mutually exclusive AU pairs, respectively. Here, $d_{i,j} = \| \nabla_{r_{m}[t]} \text{log}q_t(r_{m}[t])_i - \nabla_{r_{m}[t]} \text{log}q_t(r_{m}[t])_j \|$ represents the difference between the score functions of two distinct expression coefficients, quantifying the relationship between two individual facial action units in real faces. Similarly, $\hat{d}_{i,j} = \| p_\theta(r_{m}[t], c)_i - p_\theta(r_m[t], c)_j \|$ denotes the difference between two estimated scores, representing the facial action unit relationship estimated by the learned model. All defined AU pairs are presented in the Appendix~\ref{subsec:details_facial_action_unit_pairs}.

\begin{figure*}[t!]
    \centering
    \includegraphics[width=1.8\columnwidth]{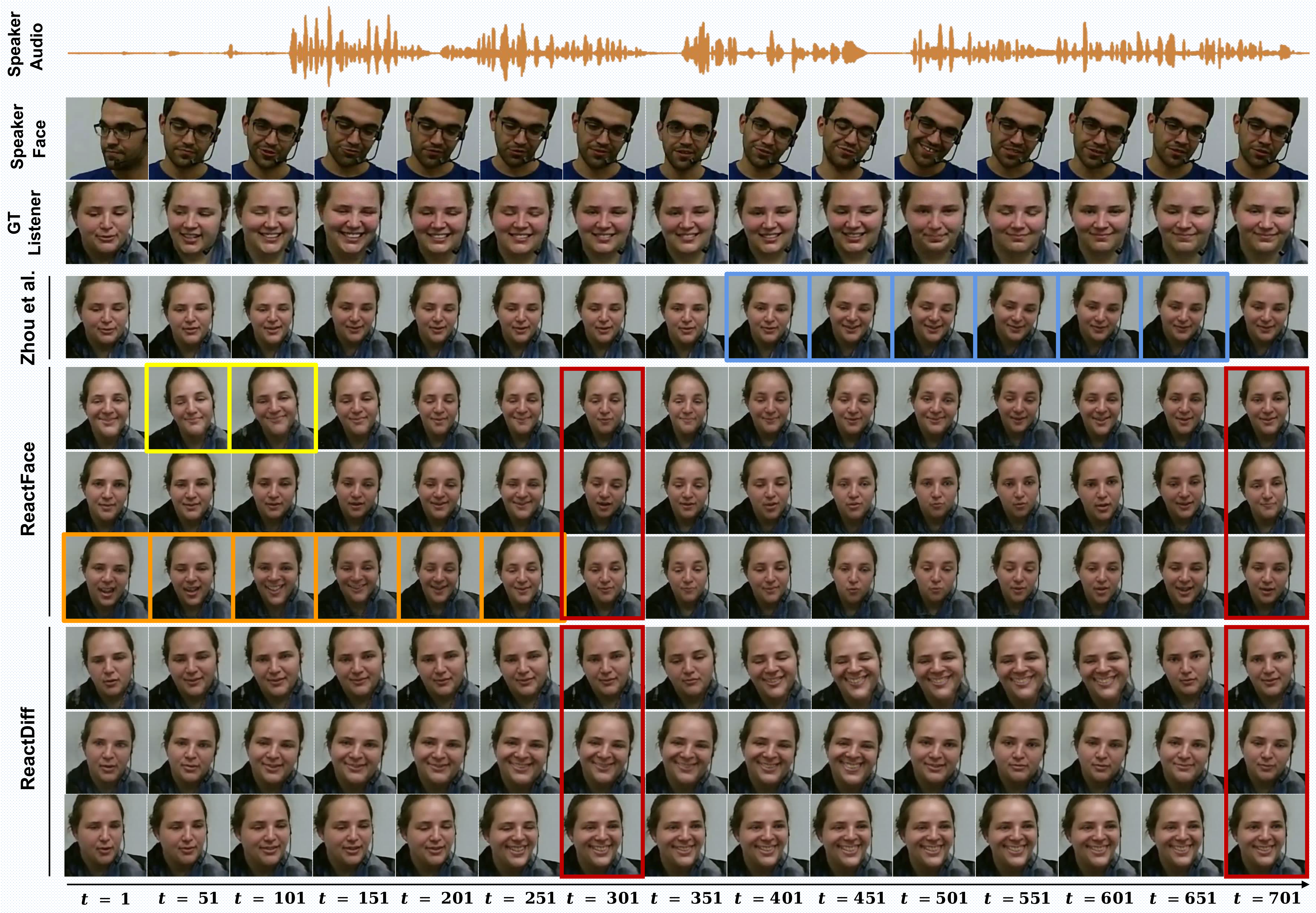} 
            \caption{\textbf{Qualitative Results} on the REACT2024  test set. Each approach generates reaction sequences online based on a given sequence of speaker visual-audio behavior. Diversity in reactions is emphasized using  red boxes, segments displaying a slow change speed are marked with  blue boxes, while those with a  rapid change speed are highlighted in  orange boxes. Frames showing unnatural facial expressions or distortions are indicated by yellow boxes.}
    \label{fig:vis_diff_sample}
\end{figure*}

\subsection{Training and Sampling}

We propose to train our ReactDiff in an simple end-to-end manner with three loss terms as:
\begin{equation}
 \mathcal{L} = \mathcal{L}_\text{dm} +  \mathcal{L}_\text{fbk} +  \lambda \mathcal{L}_\text{fac}
\label{eq:training_loss}
\end{equation}
where $\lambda$ decides the relative importance of the facial action constraint.
For sampling, we use an SDE-based solver, which is outlined and detailed in our Appendix~\ref{sec:ode_sde}. We will demonstrates the strengths of the SDE-based solver compared to the ODE-based solver in our ablation studies.

\section{Experiments}
\label{experiments}

\subsection{Experimental Setup}

\noindent \textbf{Datasets:} We evaluate the ReactDiff on a open-source hybrid video conference dataset provided by REACT2023/REACT2024 challenge \footnote{\url{https://sites.google.com/cam.ac.uk/react2024/home}} and used by previous studies \cite{luo2024reactface,xu2023reversible}, which is made up of 2962 dyadic interaction sessions (1594 in training set, 562 in validation set and 806 in test set) comes from two video conference datasets: RECOLA \cite{ringeval2013introducing} and NOXI \cite{cafaro2017noxi}, where each session contains a pair of 30s long audio-visual clips describing two subjects' interactions. 

\noindent  \textbf{Implementation details:} Our ReactDiff is trained using an AdamW optimizer \cite{kingma2014adam} with a fixed learning rate of $1e^{-4}$, $\beta_1 = 0.95$ and $\beta_2 = 0.999$ and a weight decay of $1e^{-3}$. The batch size and hyper-parameter $\lambda$ for weighting the contribution of facial action constraint $\mathcal{L}_\text{fac}$ are set to $100$ and $1e^{-4}$, respectively. Our code is implemented in PyTorch \cite{paszke2019pytorch} platform using a single Tesla A100 GPU with 40G memory and runs for total 30,000 steps for training. Our model uses 50 diffusion steps with classifier-free guidance. We follow a previous study \cite{luo2024reactface} to use the state-of-the-art 3DMM FaceVerse \cite{wang2022faceverse} to estimate the facial pose and expression coefficients, where each coefficient corresponds to an ARKit blendshape, which has an explicitly and human interpretable definition such as 'BrowInnerUp', 'EyeLookDownRight ', ' JawOpen', ' MouthFunnel', 'NoseSneerRight' and 'TongueOut'. Furthermore,  we use the PIRender \cite{ren2021pirenderer} to translate the predicted 3DMM coefficients to 2D facial reaction images. 
More details are provided in the Appendix~\ref{sec:details_configuration}.

\noindent  \textbf{Evaluation metrics:} We follow the evaluation protocol in previous works \cite{luo2024reactface,song2023multiple,song2023react2023} to assess four key aspects of the generated facial reactions: diversity, realism, appropriateness and synchrony. To evaluate diversity, we utilize three metrics: FRDvs 
to quantify diversity across reactions conditioned on different speaker behaviors, FRVar to measure variations between frames in each reaction sequence, and FRDiv to assess diversity conditioned on the given behaviors. For realism, we adopt the FVD (Fr\'echet Video Distance) \cite{unterthiner2018towards} to measure the distribution distance between generated and GT reaction sequences. We use FRCorr and FRSyn (TLCC) to evaluate the appropriateness and synchrony, respectively.

\begin{table*}[t]\centering
 \caption{
 \textbf{Quantitative Results} on REACT2024 test set. The \textbf{best} and \underline{second best} results in each column are marked in bold and underlined font, respectively.} 
 \resizebox{0.78\textwidth}{!}{
\begin{tabular}{lcccccc}
\toprule 
\multirow{2}{*}{Method} & \multicolumn{3}{c}{Diversity}  & \multicolumn{1}{c}{Realism} & \multicolumn{1}{c}{Appropriateness}  &   \multicolumn{1}{c}{Synchrony}\\
\cmidrule(r){2-4} \cmidrule(r){5-5}  \cmidrule(r){6-6}  \cmidrule(r){7-7} 
    &  FRDvs ($\uparrow$) &  FRVar ($\uparrow$)  &  FRDiv 
 ($\uparrow$) & FVD ($\downarrow$)   & FRCorr ($\uparrow$) & FRSyn ($\downarrow$) \\ \midrule

     GT   & 0.0374 & 0.0120 & 
  -   &   282.03
    &  9.480 & 
  48.46  \\
    Mirror   & 0.0374 & 0.0120 & 
  0   &  282.03
    & 0.936 & 
  42.65  \\
      Random   & 0.0415 &0.0202 & 
  0.0414   &  477.49
    & 0.127 & 
  45.82  \\
       NN motion   & 0.0420 &0.0199 & 
  0   &  452.38 
    & 0.334 & 
  46.90  \\
     NN audio   & 0.0464 & 0.0218 & 
  0  &   496.25
    & 0.017 & 
  47.67  \\
  \midrule

    Trans-AE \cite{luo2024reactface, bank2023autoencoders}  & 0.0063 & 0.0003 & 
  0   &  599.35
  &  \underline{0.245} & 
  45.01 \\
  
Ng et al. \cite{ng2022learning}  & 0.0079 &  0.0042 & 
  0.0003  & 691.24  & 0.059  & 
  45.70  \\

Zhou et al. \cite{zhou2022responsive}  &0.0106 & 0.0039  & 
  0   &527.47  & 0.104 & 45.24 \\

ReactFace \cite{luo2024reactface}
&\underline{0.0409} & \underline{0.0159}  &0.0395
&\underline{424.46}  &0.197 
 &\textbf{43.94}   \\
 
Diffusion model   & 0.0282  &0.0134   & \underline{0.0524} & 460.99
 &0.145  &45.96
   \\ \midrule \rowcolor{Gray}

ReactDiff
&\textbf{0.0594} & \textbf{0.0199} &\textbf{0.1554}  
&\textbf{386.16}  & \textbf{0.515}
 &\underline{44.56}   \\

\bottomrule
\end{tabular}
}

\label{tb:quantitative} 
\end{table*}

\subsection{Qualitative Results}
\label{subsec:Qualitative Results}
In this section, we compare qualitative results achieved by different methods for generating facial reactions in dyadic interactions. 
We specifically present key frames from a sequence predicted online in Fig.~\ref{fig:vis_diff_sample}. To assess the diversity of these predictions in response to identical speaker behavior, we employ each generation method to produce reaction samples. These samples are then displayed in adjacent rows for comparative analysis.

LSTM-based model (\emph{i.e.,} Zhou et al. \cite{zhou2022responsive}) yields deterministic results,  with the different video samples displaying identical reaction patterns so that we only present one row of results.
We can observe that the facial expressions in this sample sequence change at an extremely slow pace, failing to match the natural rhythm of human facial movements. 
Conversely, the VAE-based model with temporal enhancement (\emph{i.e.,} ReactFace \cite{luo2024reactface}) demonstrates prompt facial changes in response to the speaker. However, 
ReactFace tends to produce similar expressions and head poses, which can be observed on three adjacent rows of frames. Apart from that, some reaction segments generated by ReactFace show rapid facial movements not typically observed in natural human behavior. 
In contrast, our ReactDiff produces distinct results with more natural expressions (smiles, disgust, gazes) and less identity change or face distortion. The pace of facial movements aligns with that of GT listener reactions, neither as slow as Zhou et al. nor as fast as ReactFace. The middle and end frames in the red boxes demonstrate ReactDiff's ability to sample more diverse reactions with varying poses, expressions (\emph{e.g.,} lips, gazes) compared to the other approaches.

\subsection{Quantitative Results}
\label{subsec:Quantitative_Results}

We summarize the quantitative results on the REACT2024 test set in Tab.~\ref{tb:quantitative}. The results on the ViCo dataset  \cite{zhou2022responsive} are also provided in the Appendix. Besides the state-of-the-art methods for comparison, we also display five baselines: i) GT represents the ground-truth listener reactions; ii) Mirror refers to the visual motions of the speaker; iii) Random denotes reactions sampled from Gaussian distributions; iv) NN motion means searching the nearest neighbor (NN) of the current speaker motion segment and returning the corresponding listener segment, a commonly used synthesis method in graphics; and v) NN audio signifies searching the NN through the speaker's auditory signals.
As shown, our proposed ReactDiff method outperforms all state-of-the-art approaches in diversity across generated reactions given different conditions (FRDvs), diversity within frames (FRVar), diversity in generated reactions for the same condition (FRDiv), realism of reaction sequences (FVD), and reaction appropriateness (FRCorr).
ReactDiff achieves substantial improvements in diversity (FRDiv), realism (FVD), and appropriateness (FRCorr) compared to the second-best competitor.
We also provide results for a vanilla diffusion model baseline. In comparison, our ReactDiff, which incorporates temporal information and spatio-temporal facial kinematics, achieves superior results across all evaluation aspects (diversity, realism, appropriateness, and synchrony).

\subsection{Ablation Studies}
\label{subsec:ablation}
We  conduct five ablation studies to evaluate the effectiveness of our designed temporal index $h$ in Eq.~\ref{eq:dis} for the diffusion model, input modalities, losses, stochasticity modelling by SDE, and our selection of the number diffusion steps, respectively.

\noindent \textbf{Effectiveness of temporal index $h$.}
Without the temporal index $h$, the generation of reactions lacks awareness of the global timeline. Consequently, the resulting sequence involves
disordered changes and often contains repeated segments. 
However, all generated sequences tend to show similar jitters and repeated patterns. This similarity leads to low diversity across different sequence samples (FRDiv).
As the model without $h$ is unaware of the timestamp in the ongoing dialogue, it cannot produce long reaction sequences with high appropriateness (FRCorr) realism (FVD), and synchrony (FRSyn).

\begin{table}[t] 
\centering
 \caption{
\textbf{Ablation study}  on temporal index $h$.}
 \resizebox{0.37\textwidth}{!}{
\begin{tabular}{c|cccc}
\toprule
  & FRDiv & FVD & FRCorr  & FRSyn \\ \midrule
 w/o $h$    & 0.1064   &427.24  &0.327  &45.55 \\  \rowcolor{Gray}
  w/ $h$      & \textbf{0.1554}   &\textbf{386.16}  &\textbf{0.515}  &\textbf{44.56}\\  
\bottomrule
\end{tabular}
}

\label{tb:compare_temporal} 

\end{table}

\begin{table}[t]
\centering
 \caption{
\textbf{Ablation study} on speaker modalities.} 
 \resizebox{0.43\textwidth}{!}{
\begin{tabular}{cc|ccccc}
\toprule
Face & Audio    &FRVar & FRDiv & FVD & FRCorr  & FRSyn \\ \midrule
    &          &0.0211   &0.0963   &467.53  & 0.048  &46.31 \\ 
    \checkmark  &  &\textbf{0.0293}     &0.1205    &442.04   &0.121  &44.94  \\

      &   \checkmark       &0.0210      &0.1028    &419.14   &0.075  &46.40  \\  \rowcolor{Gray}
         \checkmark      &   \checkmark       & 0.0199   & \textbf{0.1554}   &\textbf{386.16}  &\textbf{0.515}  &\textbf{44.56} \\  

\bottomrule
\end{tabular}
}

\label{tb:ab_modalities} 

\end{table}

\begin{table}[t]
\centering
 \caption{
\textbf{Ablation study} on two proposed constraints.} 
 \resizebox{0.48\textwidth}{!}{
\begin{tabular}{cc|ccccc}
\toprule
$\phi_{\text{FBK}}(\cdot)$ & $\phi_{\text{FAC}}(\cdot)$   &FRDvs & FRDiv & FVD & FRCorr  & FRSyn \\ \midrule
    &         &\textbf{0.1069}   &0.0996   &  425.49 &0.369  &44.98  \\ 
    
    \checkmark  &        & 0.0393 &0.0682   &383.66  &0.477  &44.97   \\

      &   \checkmark  &0.0695   & 0.1142 &\textbf{334.52}  & 0.474
& 44.74  \\  \rowcolor{Gray}
         \checkmark      &   \checkmark            & 0.0594   & \textbf{0.1554}   &386.16  &\textbf{0.515}  &\textbf{44.56} \\

\bottomrule
\end{tabular}
}

\label{tb:ab_losses} 

\end{table}

\begin{figure}[t!]
    \centering
    \includegraphics[width=1\columnwidth]{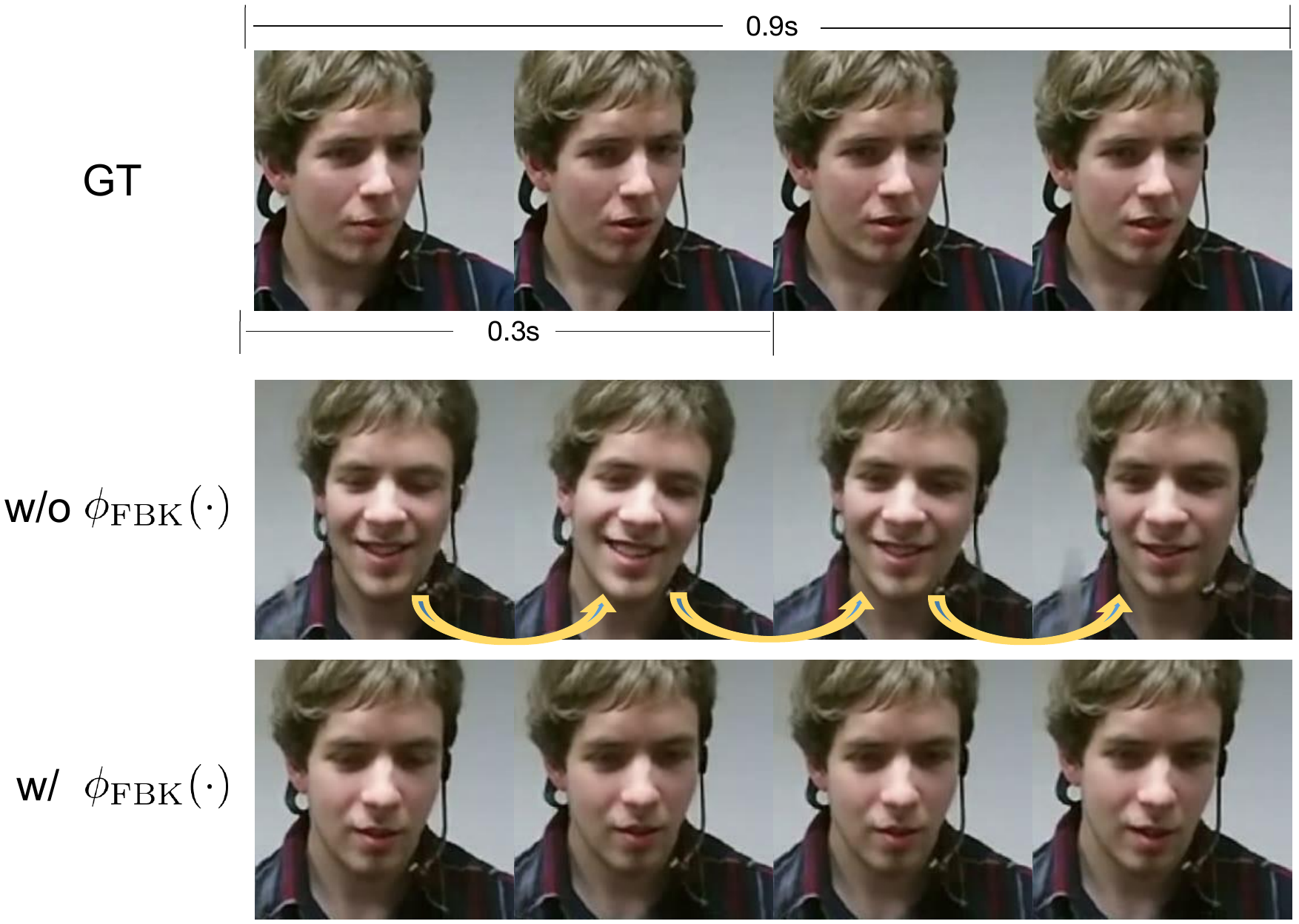} 
            \caption{Comparison of reactions from model without (w/o) the human temporal facial behavioral kinematics constraint $\phi_{\text{FBK}}(\cdot)$ and those from model with (w/) $\phi_{\text{FBK}}(\cdot)$. }
    \label{fig:vis_analysis}
\end{figure}

\noindent \textbf{Effectiveness of modalities of speaker behavior.} 
The results in  Tab.~\ref{tb:ab_modalities} show that each 
modality of speaker behavior contributes to the reaction generation. Especially, visual signals play a crucial role in improving appropriateness and synchrony of reactions, and auditory signals influence more on the realism.
With all input modalities combined, our model achieves the best performance in realism (FVD), appropriateness (FRCorr), and synchrony (FRSyn), demonstrating the complementary nature of each modality. The audio modalities constrain trajectory variations within sequences, aligning facial reactions with the rhythm of speaker behavior (such as speech content and prosody) and reducing random changes. This constraint significantly contributes to the improvements in appropriateness.

\noindent \textbf{Effectiveness of proposed losses.} 
Tab.~\ref{tb:ab_losses} shows the comparison of ReactDiff and its variants without human temporal facial behavioral kinematics constraint $\phi_{\text{FBK}}(\cdot)$ or facial spatial action constraint $\phi_{\text{FAC}}(\cdot)$.  For variant without $\phi_{\text{FBK}}(\cdot)$, the diversity within frames (FRDvs) increases due to jitters and unsmooth transitions, whereas appropriateness (FRCorr) decreases.
Fig.~\ref{fig:vis_analysis} shows that the variant without 
$\phi_{\text{FBK}}(\cdot)$ produces reactions with abrupt changes. 
For variant without $\phi_{\text{FBK}}(\cdot)$, the appropriateness decreases as more unnatural expressions appear in sequences.

\begin{figure}[t!]
    \centering
    \includegraphics[width=1\columnwidth]{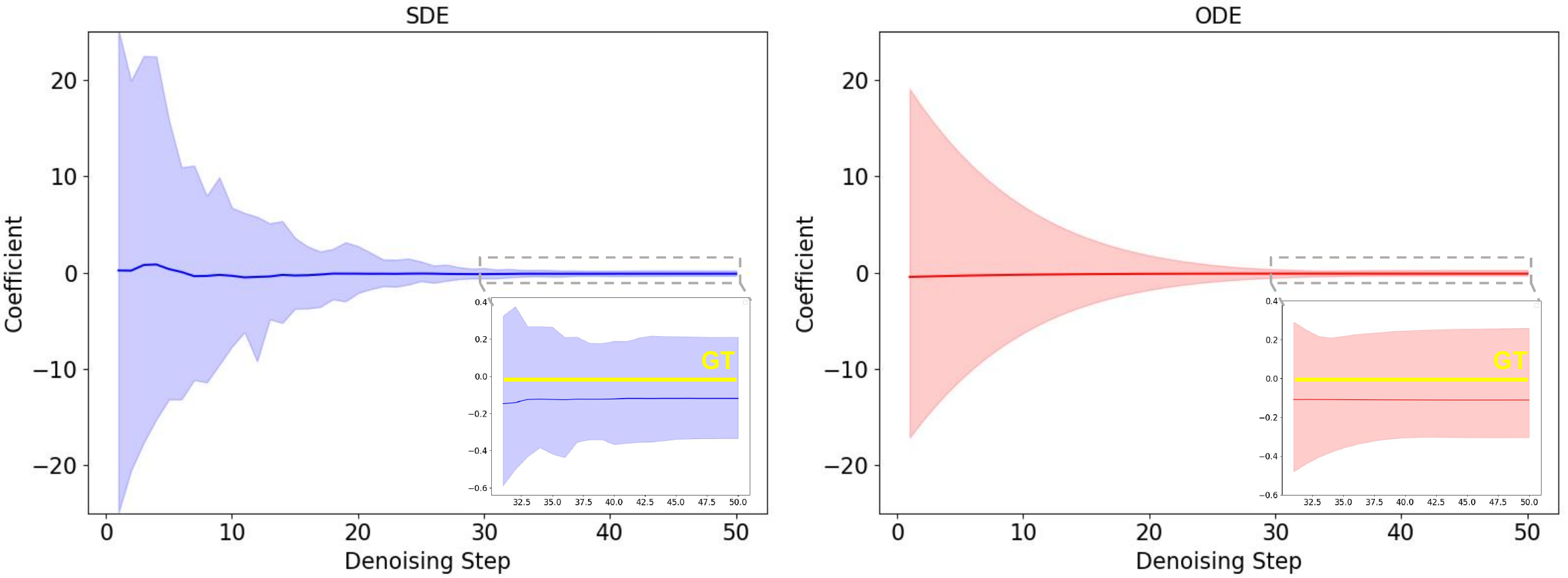} 
            \caption{Evolution of mean coefficient in diffusion denoising steps:  SDE solver \emph{vs.} ODE solver. }
\label{fig:compare_SDE-ODE}
\end{figure}

\begin{table}[t]
 \centering
  \caption{
Comparison of SDE and ODE.}
 \resizebox{0.48\textwidth}{!}{
\begin{tabular}{c|ccccc}
\toprule
Sampling    &FRVar & FRDiv & FVD & FRCorr  & FRSyn \\ \midrule
  ODE          &0.0119   &0.0857 &421.53  &0.447 &44.91  \\  \rowcolor{Gray}
   SDE         & \textbf{0.0199}   &\textbf{0.1554}   &\textbf{386.16}  &\textbf{0.515}  &\textbf{44.56}\\  
\bottomrule
\end{tabular}
}

\label{tb:compare_sde_ode} 

\end{table}
\noindent \textbf{Effectiveness of stochasticity modelling by SDE.}  
To analyze the contribution of using a SDE solver that injects independent noise (as a standard Wiener process term) at each denoising step, we compare sampling with a SDE solver versus sampling with an ODE (without a Wiener process term) solver. 
Fig.~\ref{fig:compare_SDE-ODE} shows the evolution of the mean 3DMM coefficients over denoising steps. We observe that the SDE solver obtains denoised samples in a more stochastic and wider range compared to the ODE solver, however, these samples still approach an appropriate distribution. 
The results in Tab.~\ref{tb:compare_sde_ode} also show that sampling using the SDE solver achieves superior diversity (FRVar and FRDiv). 
Despite the SDE injects more stochasticity, it can also achieve higher appropriateness. The reason is that the generated reactions resemble human-like variability rather than converging to an averaged mode of behavior.

\noindent \textbf{Analysis of denoising steps.}  
Tab.~\ref{tb:compare_denoising_steps} presents the results sampled with different denoising steps. We found that denoising with fewer steps leads to less diversity and an averaging mode of samples, although with high appropriateness. Finally, we choose 50 steps as our setting.

\begin{table}[t]
\centering
 \caption{
The influence of denoising steps.} 
\resizebox{0.42\textwidth}{!}{
\begin{tabular}{c|cccccc}
\toprule
Step   &FRVar & FRDiv & FVD & FRCorr  & FRSyn  \\ \midrule
    2         &0.0023   &0.0092  &560.74  &\textbf{0.515}  &\textbf{44.43}   \\ 
   5        &0.0176   & 0.1003   &432.41  &0.460  &44.77  \\
   10     &\textbf{0.0203}   &0.1059   &410.42  &0.451  &44.93  \\ 
   25         &0.0199   &0.1553   &421.08  &\textbf{0.515}  &44.57  \\   \rowcolor{Gray}

   50    & 0.0199   & \textbf{0.1554}   &\textbf{386.16}  &\textbf{0.515}  &44.56   \\  
   100         &0.0171   &0.0791   &415.01  &0.497  &44.70  \\  
\bottomrule
\end{tabular}
}

\label{tb:compare_denoising_steps} 
\end{table}

\section{Conclusion}
\label{conclusion}

We have proposed ReactDiff, a novel diffusion model for online generation of multiple appropriate facial reactions in dyadic interactions. By introducing temporal modeling and spatio-temporal facial kinematics priors into the diffusion denoising process, we enable model to generate a set of human-like reaction samples, effectively avoiding artifacts such as jitters, abrupt transitions, and repeated segments. Experiments demonstrate ReactDiff's superior performance in producing diverse, appropriate, and realistic reactions in response to speakers.

\bibliographystyle{ACM-Reference-Format}
\bibliography{sample-base}

\clearpage
\appendix

\section{Details of ReactDiff}
\label{sec:details_configuration}
We present the hyperparameter details in Tab.~\ref{tb:hyperparam_details}. For our diffusion model network, we opted for a UNet architecture. The model underwent training utilizing the AdamW optimizer \cite{kingma2014adam} with a consistent learning rate set at $1 \times 10^{-4}$, $\beta_1 = 0.95$, $\beta_2 = 0.999$, and a weight decay of $1 \times 10^{-3}$.
The batch size was set to $100$, while the hyperparameter $\lambda$, responsible for weighing the contribution of the facial action constraint $\mathcal{L}_{fac}$, was set to $1 \times 10^{-4}$.
Our model operates with 50 diffusion steps employing a classifier-free guidance approach. The strategy for our noise levels sampling aligns with previous methodologies described in the work of Karras et al. 
\cite{karras2022elucidating}.
A state-of-the-art pre-trained wav2vec2.0 speech model \cite{baevski2020wav2vec} is leveraged to encode the raw audio signal as a set of speaker auditory embeddings.
\begin{table}[h]
\caption{
Hyperparameters.} 
\centering
 \resizebox{0.4\textwidth}{!}{
\begin{tabular}{l|c}
\toprule
Parameter   &  Value \\ \midrule
  Batch size       & 100\\ 
  Num. of diffusion steps    & 50\\ 
  Num. of training iterations    & 30k \\
  Noise Schedule & Cosine \\
  Window size $w$ & 16 \\
  Audio Encoder &  Wav2Vec2.0  \\ 
   Optimizer & AdamW   \\ 
  Learning rate     & $1.0 \times 10^{-4}$ \\ 
  Weight decay  & $1.0 \times 10^{-3}$   \\ 
     Weighting $\lambda$ for facial action constraint        &  $1.0 \times 10^{-4}$    \\  

  $\beta_1$    &  0.95  \\ 

   $\beta_2$    &  0.999 \\

\bottomrule
\end{tabular}
}

\label{tb:hyperparam_details} 

\end{table}

\subsection{Conditioned Generation}
\label{sec:network_details_supp}
\begin{figure*}[h!]
    \centering
    \includegraphics[width=2\columnwidth]{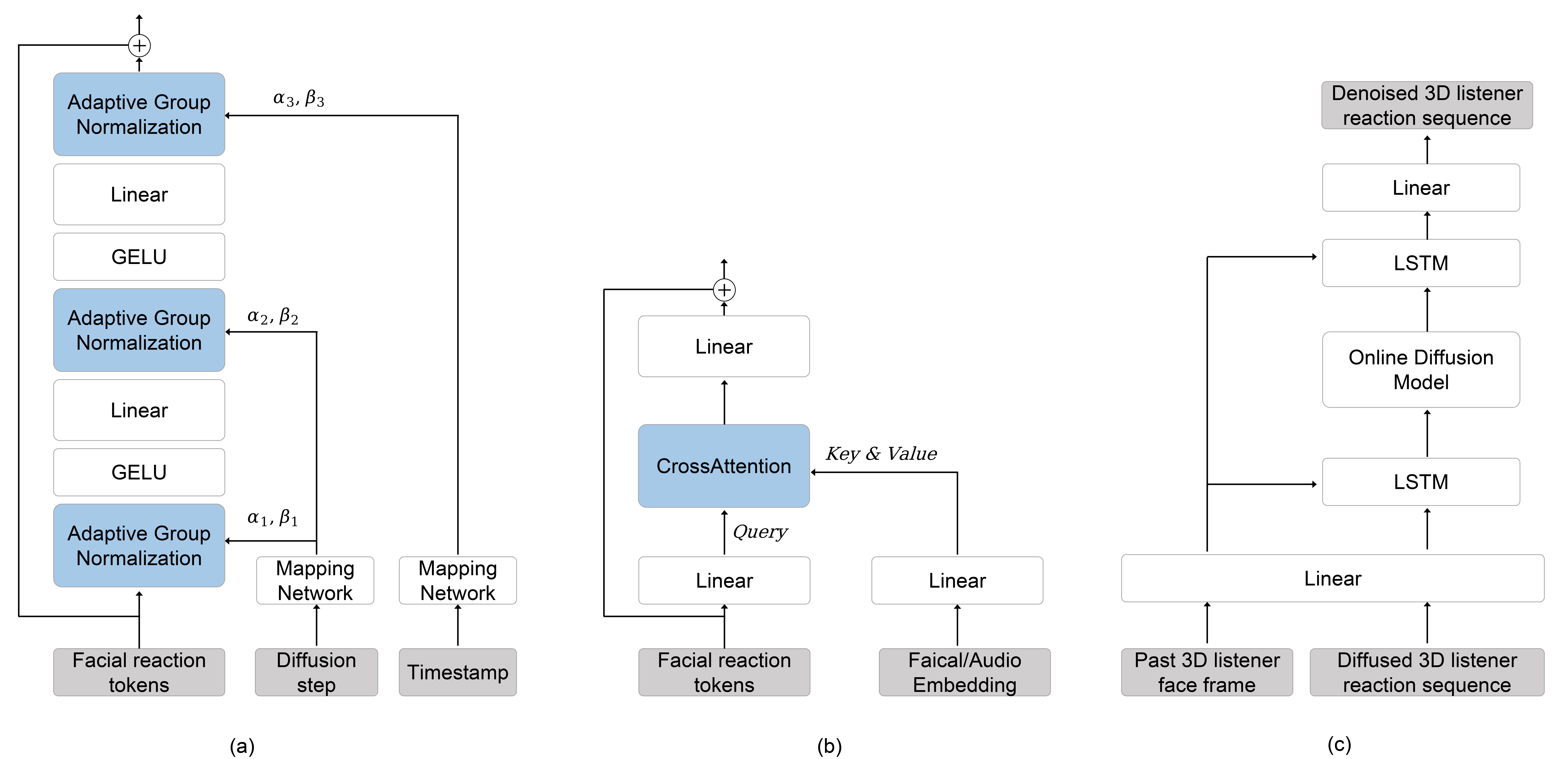} 
        \caption{Condition incorporation through (a) Adaptive Group Normalization in ResBlock, (b) Cross-Attention in CrossAttentionBlock, and (c) LSTM in input and output layers}
    \label{fig:codition}
\end{figure*}

This section provides a more comprehensive overview of the condition incorporation used in our architecture.
We used adaptive group normalization to incorporate the diffusion step condition and timestamp (a global timestamp in an ongoing conversation), as shown in Fig.~\ref{fig:codition} (a). This method allows the model to adjust its normalization parameters dynamically based on the diffusion step, enhancing its adaptability and performance across different stages of the diffusion process.
For conditioning facial reaction sequences on the speaker's facial and auditory sequences, we employed cross-attention mechanisms, as shown in Fig.~\ref{fig:codition} (b). In the CrossAttentionBlock, the speaker's conditions, comprising both facial expressions and audio features, are utilized as keys and values, while the listener's facial reaction sequences serve as queries. This approach enables the model to effectively integrate contextual information from the speaker, ensuring that the listener's reactions are appropriately synchronized with the speaker's cues.
To prevent previous tokens from accessing information from future tokens, we incorporated causal masks in the attention operations. This ensures that the attention mechanism adheres to the temporal sequence of the data, preserving the chronological order of events and maintaining the integrity of the sequence prediction.
For conditioning facial reaction sequences on historical information, we employed an one-layer LSTM before and after generation process of online diffusion model, as shown in Fig.~\ref{fig:codition} (c).
Specially, we used past 3D listener face frame as the initialized hidden state in LSTMs.

\label{sec:conditions}

\section{ODE and SDE Solvers}
\label{sec:ode_sde}
We provide a comprehensive overview of the SDE and ODE solvers utilized in our methodology in Algorithm~\ref{alg:dpm-solver++m-ode} and Algorithm~\ref{alg:dpm-solver++m-sde}, respectively, and highlight their distinctions.
Specifically, Algorithm~\ref{alg:dpm-solver++m-ode} illustrates the ODE variant of the DPM-Solver++(2M), a second-order multistep solver introduced in prior research \cite{lu2022dpm}. Conversely, Algorithm~\ref{alg:dpm-solver++m-sde} elucidates the SDE counterpart of the solver, showcasing the differential equation-based approach. 
This comparative outline emphasizes the differential aspects and distinctive operational mechanisms between the ODE and SDE solvers.

\begin{algorithm}[th]
    \centering
    \caption{DPM-Solver++ 2M (ODE).}\label{alg:dpm-solver++m-ode}
    \begin{algorithmic}[1]
    \Require initial value $\boldsymbol{x}_T$, time steps $\{t_i\}_{i=0}^T$, noise levels $\{\sigma_i\}_{i=0}^T$,  score matching network $\boldsymbol{s}_\theta$.
        \State Denote $h_i \coloneqq \lambda_{t_{i}} - \lambda_{t_{i-1}}$, 
        for $i=1,\dots,T$.
        \State $\tilde{\boldsymbol{x}}_{t_0} \gets \boldsymbol{x}_T$.

        \State $\tilde{\boldsymbol{x}}_{t_1} \gets \frac{\sigma_{t_{1}}}{\sigma_{t_0}}\tilde{\boldsymbol{x}}_{t_0} - \left(e^{-h_1} - 1\right)s_\theta(\tilde{\boldsymbol{x}}_{t_0}, t_0)$

        \For{$i\gets 2$ to $T$}
        \State $r_i \gets \frac{h_{i-1}}{h_{i}}$
        \State $\boldsymbol{D}_i \leftarrow \left(1+\frac{1}{2 r_i}\right) \boldsymbol{s}_\theta\left(\tilde{\boldsymbol{x}}_{t_{i-1}}, t_{i-1}\right) -\frac{\boldsymbol{s}_\theta(\tilde{\boldsymbol{x}}_{t_{i-2}}, t_{i-2})}{2 r_i}  $
   
    \State  $\tilde{\boldsymbol{x}}_{t_{i}} \gets \frac{\sigma_{t_{i}}}{\sigma_{t_{i-1}}} \tilde{\boldsymbol{x}}_{t_{i-1}} - \left(e^{-h_i} - 1\right)D_i$        
        \EndFor
        \State \Return $\tilde{\boldsymbol{x}}_{t_T}$
    \end{algorithmic}
\end{algorithm}

\begin{algorithm}[th]
    \centering
    \caption{DPM-Solver++ 2M (SDE).}\label{alg:dpm-solver++m-sde}
    \begin{algorithmic}[1]
    \Require initial value $\boldsymbol{x}_T$, time steps $\{t_i\}_{i=0}^T$, noise levels $\{\sigma_i\}_{i=0}^T$,  score matching network $\boldsymbol{s}_\theta$, $\eta$.
        \State Denote $h_i \coloneqq \lambda_{t_{i}} - \lambda_{t_{i-1}}$ 
        for $i=1,\dots,T$.
        
        \State $\tilde{\boldsymbol{x}}_{t_0} \gets \boldsymbol{x}_T$. 

        \State $\tilde{\boldsymbol{x}}_{t_1} \gets
        e^{-\eta h_1}
        \frac{\sigma_{t_{1}}}{\sigma_{t_0}} 
        \tilde{\boldsymbol{x}}_{t_0} - (e^{-h_1 - \eta h_1 } - 1) s_{\theta}(\tilde{\boldsymbol{x}}_{t_0}, t_0) + \sigma_{t_{1}} \sqrt{1-e^{-2\eta h_1}} z_{t_{1}} $

        \For{$i\gets 2$ to $T$}
        \State $r_i \gets \frac{h_{i-1}}{h_{i}}$

        \State $\boldsymbol{D}_i \gets
        \left(1+\frac{1}{2 r_i}\right)
                s_{\theta}(\tilde{\boldsymbol{x}}_{t_{i-1}}, t_{i-1})  - \frac{ s_{\theta}(\tilde{\boldsymbol{x}}_{t_{i-2}}, t_{i-2})}{2 r_i}$

        \State $\boldsymbol{N}_i \gets
     \sigma_{t_{i}} \sqrt{1-e^{-2\eta h_i}} z_{t_{i}}$

        \State $\tilde{\boldsymbol{x}}_{t_i} \gets
        e^{-\eta h_i}
        \frac{\sigma_{t_{i}}}{\sigma_{t_{i-1}}} 
        \tilde{\boldsymbol{x}}_{i-1} 
       - (e^{(-h_i - \eta h_i )}-1) \boldsymbol{D}_i + \boldsymbol{N}_i$

        \EndFor
        \State \Return $\tilde{\boldsymbol{x}}_{t_T}$
    \end{algorithmic}
    
\end{algorithm}

where the variable $\lambda_t = \text{log}(\alpha_t / \sigma_t)$ signifies the logarithm of the Signal-to-Noise Ratio (SNR) and is a strictly decreasing function of $t$, the noise term $z_{t_i} \sim \mathcal{N}(\mathrm{0}, \mathrm{I})$ follows a Gaussian distribution with zero mean and identity covariance. Here, $\alpha_t$ denotes the mean and $\sigma_t$ represents the standard deviation of the noise distribution at level $t$. For a comprehensive understanding of these concepts and details, please refer to the work by Lu et al. \cite{lu2022dpm}.

Upon comparing the characteristics of two algorithms, it becomes apparent that the SDE solver incorporates an additional component, denoted as $\boldsymbol{N}_i$, which includes stochastic factors  in the reverse-time diffusion process. This augmentation presents a notable distinction between the methodologies under consideration.

\begin{table*}[t]\centering
 \caption{
 \textbf{Quantitative Results of Expression Coefficients} on REACT2024 test set. The \textbf{best} and \underline{second best} results in each column are marked in bold and underlined font, respectively.} 
 \resizebox{0.7\textwidth}{!}{
\begin{tabular}{lcccccc}
\toprule 
\multirow{2}{*}{Method} & \multicolumn{3}{c}{Diversity}    & \multicolumn{1}{c}{Appropriateness}  &   \multicolumn{1}{c}{Synchrony}\\
\cmidrule(r){2-4} \cmidrule(r){5-5}  \cmidrule(r){6-6}  
    &  FRDvs ($\uparrow$) &  FRVar ($\uparrow$)  &  FRDiv 
 ($\uparrow$)   & FRCorr ($\uparrow$) & FRSyn ($\downarrow$) \\ \midrule

     GT   & 0.0330 & 0.0104 & 
  -  
    &  9.424 & 
48.49
  
  \\
    Mirror  & 0.0330 & 0.0104 & 
  -   
    &  9.424 & 
  42.65 
  
  \\
      Random   & 0.0348 &0.0169 & 
 0.0348   
    & 0.132 & 
  45.88  \\

       NN motion   & 0.0348 & 0.0164 & 
  0  
    & 0.327 & 
  46.85

  \\
     NN audio   & 0.0408 & 0.0192 & 
  0  
    & 0.154 & 
  47.68 \\
  \midrule

    Trans-AE \cite{luo2023reactface}  & 0.0003 & 0.0001 & 
  0  
  &  0.046 & 
  44.91 \\

Ng et al. \cite{ng2022learning}  & 0.0001 &  0.0003 & 
  0.0001   &0.091  & 
  46.07 \\

Zhou et al. \cite{zhou2022responsive}  &0.0006 & 0.0002 & 
  0    &0.021 & 46.94\\

ReactFace \cite{luo2023reactface}
&0.0017 & 0.0007  &-
  &0.103
 &\textbf{44.51}  \\

Diffusion model   & \underline{0.0274}  &\underline{0.0131}   & \underline{ 0.0510} 
 &\underline{0.265} &\underline{44.83}
   \\ \midrule \rowcolor{Gray}

ReactDiff
&\textbf{0.1285} & \textbf{0.0527} &\textbf{0.2175}  
  & \textbf{0.403}
 &45.45   \\

\bottomrule
\end{tabular}
}

\label{tb:quantitative_exp} 
\end{table*}

\begin{table*}[t]\centering
 \caption{
 \textbf{Quantitative Results of Pose Coefficients} on REACT2024 test set. The \textbf{best} and \underline{second best} results in each column are marked in bold and underlined font, respectively.} 
 \resizebox{0.7\textwidth}{!}{
\begin{tabular}{lcccccc}
\toprule 
\multirow{2}{*}{Method} & \multicolumn{3}{c}{Diversity}    & \multicolumn{1}{c}{Appropriateness}  &   \multicolumn{1}{c}{Synchrony}\\
\cmidrule(r){2-4} \cmidrule(r){5-5}  \cmidrule(r){6-6}  
    &  FRDvs ($\uparrow$) &  FRVar ($\uparrow$)  &  FRDiv 
 ($\uparrow$)   & FRCorr ($\uparrow$) & FRSyn ($\downarrow$) \\ \midrule

   GT &  0.0761 & 0.0267 & 
  -   
    & 1.711 & 20.98
  
  \\
    Mirror  &  0.0761 & 0.0267 & 
  -   
    & 1.711 & 0
  
  \\
      Random   & 0.0992 &0.0484 & 
0.0989 
    & 0.532 & 
  27.58\\

       NN motion   & 0.1045 & 0.0501 & 
  0  
    & 0.577 & 
  20.15

  \\
     NN audio   & 0.0951 & 0.0440 & 
  0  
    & 0.057 & 
  30.35 \\
  \midrule

    Trans-AE \cite{luo2023reactface}  &  0.0022 & 0.0001 & 
  0  
  &  0.049& 
 29.69 \\

Ng et al. \cite{ng2022learning}  & 0.0007 & 0.0001 & 
  0.0001   &0.103  & 
  28.46\\

Zhou et al. \cite{zhou2022responsive}  &0.0031 &0.0023 & 
  0    &\underline{0.349} & \textbf{20.62}\\

ReactFace \cite{luo2023reactface}
&0.0009 &0.0001  &0.0395
  &0.093
 &20.92  \\

Diffusion model   & \underline{0.0360}  &\underline{0.0166}   & \underline{0.0634} 
 &0.280 &24.74
   \\ \midrule \rowcolor{Gray}

ReactDiff
&\textbf{0.0426} & \textbf{0.0022} &\textbf{0.0683}  
  & \textbf{0.463}
 &\underline{20.67}   \\

\bottomrule
\end{tabular}
}

\label{tb:quantitative_pose} 
\end{table*}

\section{Additional Results}
\subsection{Additional Results on REACT2024 Dataset}

To thoroughly assess the efficacy of ReactDiff, we present additional experimental results comparing the generated expression coefficients (shown in Table \ref{tb:quantitative_exp}) and pose coefficients (detailed in Table \ref{tb:quantitative_pose}).
Our observations reveal that ReactDiff achieves heightened diversity in both facial expressions and poses. In  comparison to other generative models, ReactDiff distinctly improves the appropriateness of generated facial expressions or poses. These findings demonstrate the effectiveness of our proposed methodology in significantly enhancing the fidelity and quality of synthesized facial expressions and poses.

\begin{table}[th]
\centering

 \caption{
Comparison of quantitative results on ViCo test set.}
\resizebox{0.45\textwidth}{!}{
 \begin{tabular}{lccccc}
 \toprule
\multicolumn{2}{l|}{Methods}   & FRVar  & FRDiv  & FVD & FRSyn\\
\midrule 

\multicolumn{2}{l|}{GT}  

 &1.8439 & - & 168.24
&29.61
 \\

\multicolumn{2}{l|}{ Trans-AE} 

 & 0.0145  &  0 & 250.09
& 32.52
 \\
\multicolumn{2}{l|}{ Ng et al. \cite{ng2022learning}} 
 &\textbf{1.1032}  &   0 & 460.48

&31.00
 \\
\multicolumn{2}{l|}{Zhou et al. \cite{zhou2022responsive}} &  0.9314
&   0 & \textbf{180.56} & 32.62   \\

\multicolumn{2}{l|}{ReactFace}
 &0.3539  &0.3015 &271.09 &31.12\\

\multicolumn{2}{l|}{ReactDiff}
 &0.5777 & \textbf{0.4074} & 188.32  & \textbf{26.01}\\
\bottomrule
 \end{tabular} 
 }
 
\label{tb:compare_vico_} 

\end{table}

\begin{table*}[t]\centering

 \caption{
 \textbf{Quantitative Results} on ViCo test set. The \textbf{best} results in each column is marked in bold.} 
 \resizebox{0.78\textwidth}{!}{
\begin{tabular}{lcccccccc}
\toprule 
\multirow{2}{*}{Method} & \multicolumn{4}{c}{Realism}   & \multicolumn{3}{c}{Feature Distance ($\downarrow$)} \\
\cmidrule(r){2-5}  \cmidrule(r){6-8}  
    &  FID ($\downarrow$) &  SSIM($\uparrow$)  &  PSNR 
 ($\uparrow$)   & CPBD ($\uparrow$) & Angle & Exp & Trans \\ \midrule

      Random  &-   & -  & -
  
    &-  &18.04  &44.67 &19.80\\

Zhou et al. \cite{zhou2022responsive}  &\textbf{30.53} &0.601 & 
  18.15    &0.126 &\textbf{7.79} &\textbf{15.04} &\textbf{6.52}\\

Diffusion model   &57.99   &\textbf{0.618}  & 17.20
  
    &0.147  &14.29  &21.65  &10.09
   \\ \midrule \rowcolor{Gray}

ReactDiff
&56.25   &0.616 & \textbf{18.19}
  
    &\textbf{0.148}  &8.68  &21.02 &9.59 \\

\bottomrule
\end{tabular}
}

\label{tb:quantitative_test} 
\end{table*}

\begin{table*}[t]\centering
 \caption{
 \textbf{Quantitative Results} on ViCo ood (out of distribution) set. The \textbf{best} results in each column is marked in bold.} 
 \resizebox{0.78\textwidth}{!}{
\begin{tabular}{lcccccccc}
\toprule 
\multirow{2}{*}{Method} & \multicolumn{4}{c}{Realism}   & \multicolumn{3}{c}{Feature Distance ($\downarrow$)} \\
\cmidrule(r){2-5}  \cmidrule(r){6-8}  
    &  FID ($\downarrow$) &  SSIM($\uparrow$)  &  PSNR 
 ($\uparrow$)   & CPBD ($\uparrow$) & Angle & Exp & Trans \\ \midrule

   Random  &-   & -  & -
  
    &- &18.11  &44.60 &20.36 \\

Zhou et al. \cite{zhou2022responsive}  &\textbf{24.96} &0.521 & 
  16.56   &\textbf{0.142} &8.23 &22.83 &8.32\\

Diffusion model  &49.89   &0.506 & 15.72
  
    &0.088  &7.66  &22.89 &8.78
   \\ \midrule \rowcolor{Gray}

ReactDiff
&47.88    &\textbf{0.543}  & \textbf{16.62}
  
    &0.083  &\textbf{7.10}  &\textbf{21.79} &\textbf{8.05} \\

\bottomrule
\end{tabular}
}

\label{tb:quantitative_ood} 
\end{table*}

\subsection{Additional Results on ViCo Dataset}
We extend our experimental analysis to further include results on the ViCo dataset \cite{zhou2022responsive}. This dataset comprises data from 92 subjects, consisting of 67 speakers and 76 listeners, with a total of 483 video clips sourced from YouTube. Notably, the ViCo dataset lacks 'appropriate facial reaction' labels.
Consequently, we can not assess the appropriateness.
The results presented in Tab.~\ref{tb:compare_vico_} indicate that our ReactDiff method achieves competitive realism, as measured by FVD (Fréchet Video Distance), and showcases superior synchronicity (FRSyn) and diversity (FRDiv).

Following the evaluation protocols established in Zhou et al.'s work \cite{zhou2022responsive}, we conducted performance evaluation of various generative methods on ViCo test and out-of-distribution (ood) sets. The evaluation employed metrics such as FID (Fréchet Inception Distance), SSIM (Structural Similarity Index), PSNR (Peak Signal-to-Noise Ratio), CPRB (Coefficient Path Rank Breakdown), and feature distance metrics. These metrics assess video quality and the proximity between the generated and ground-truth coefficients.

The results, as presented in Tab.~\ref{tb:quantitative_test} for the test set and Tab.~\ref{tb:quantitative_ood} for the out-of-distribution set, demonstrate the superiority of our ReactDiff method. Specifically, our method outperforms others in 5 out of 7 cases on the ood set and showcases competitive performance on the test set.
It is worth noting that the generative technique proposed by Zhou et al. yields deterministic reaction results characterized by a slow pace of changes, consequently resulting in lower diversity within their generated results. However, their method excels in metrics such as FID and Feature Distance, particularly in terms of proximity to ground-truth coefficients.
The observed phenomenon stems from the fact that the generated reactions closely align with the ground-truth coefficients. The deterministic nature of the method results in fewer variations in the generated reactions. Consequently, while this approach excels in accurately mapping to the ground-truth coefficients, it shows limited diversity due to its deterministic nature, leading to fewer variations in the generated outputs.

\subsection{Sensitivity Analysis}

Tab.\ref{tb:compare_hyperparam_sensitivity_analysis} illustrates the sensitivity analysis conducted on the hyperparameter $\lambda$ within the framework of the overall training loss (Eq.~\ref{eq:training_loss}). This hyperparameter plays a crucial role in determining the relative significance of the facial action constraint. The findings demonstrate that variations in the value of $\lambda$ significantly influence the appropriateness metric, resulting in a decrease from 0.515 to 0.469 for larger values (\emph{e.g.}, 1) and from 0.515 to 0.477 for smaller values (\emph{e.g.}, 0). These results indicate that an excessive facial action constraint can impede the efficacy of the diffusion training process. Conversely, an absence of such constraints results in generated facial reactions that deviate towards unnatural expressions.
In the end, a value of $\lambda = 0.0001$ was selected, deemed appropriate within the context of the study.

\begin{table}[th]
 \resizebox{0.48\textwidth}{!}{
\begin{tabular}{c|ccccc}
\toprule
$\lambda$   &FRVar & FRDiv & FVD & FRCorr  & FRSyn  \\ \midrule
   1        & 0.0117   &0.0593   &387.26  &0.469  &44.91  \\
   0.1     &0.0230   &0.0915   &373.12  &0.387  &45.15  \\ 
   0.01   &0.0330   &0.1928   &418.59  &0.437  &44.64  \\ 
  0.001   &0.0111 & 0.0631 & 472.05 & 0.556 & 43.68  \\ \rowcolor{Gray}
  0.0001       & 0.0199   & 0.1554   &386.16  &0.515  &44.56\\ 
    0.00001       & 0.0280  &0.1893   &408.65  &0.152  &44.74\\ 
   0      &0.0160  &0.0682   &440.13  &0.477  &44.97 
   \\  

\bottomrule
\end{tabular}
}
 \caption{
Sensitivity analysis of hyperparameter $\lambda$.} 
\label{tb:compare_hyperparam_sensitivity_analysis} 
\end{table}

\subsection{Perception Survey}
\label{subsec:perceptual}

\begin{table}[t!]
\centering
 \caption{\label{tb:user-study} User preference results between the facial reactions generated by our ReactDiff and competitors.
}
 \resizebox{0.48\textwidth}{!}{
 \begin{tabular}{lccccc}
 \toprule
\multicolumn{2}{l|}{\textbf{Ours \emph{vs.} Competitor}}& Realism  &Diversity &Appropriateness &Sync   \\
\midrule 

\multicolumn{2}{l|}{Ours \emph{vs.} Zhou et al. \cite{zhou2022responsive}}&71.4\%  & 100\%  & 80.9\%   &85.7\%  \\

\multicolumn{2}{l|}{Ours \emph{vs.} Ng et al. \cite{ng2022learning}}& 78.6\%  & 69.1\%  &  69.1\% & 69.1\%  \\

\multicolumn{2}{l|}{Ours \emph{vs.} ReactFace \cite{luo2023reactface}}&80.5\%  & 66.7\%  & 69.1\%   &52.4\%  \\
\multicolumn{2}{l|}{Ours \emph{vs.} GT}& 45.2\%   &59.5\%  &40.5\%  &45.2\%   \\
\bottomrule

 \end{tabular} 
 }

\label{ex:user_study}
\end{table}

\begin{figure*}[th]
    \centering
    \includegraphics[width=1.4\columnwidth]{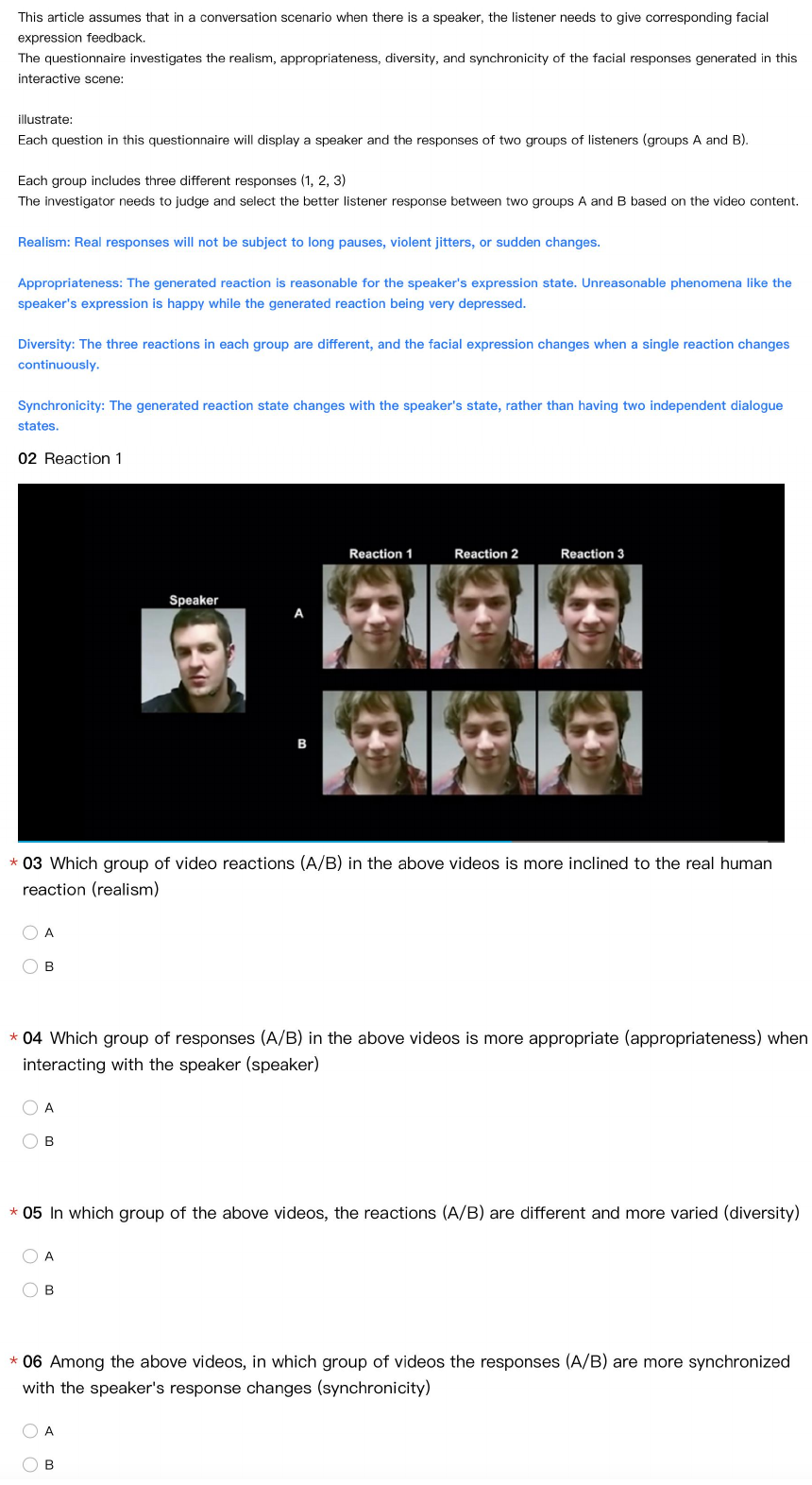} 
        \caption{Designed user interface on Tencent Questionnaire platform. Each comparison contains two groups of  generated reaction sequences.}
    \label{fig:use_study}
\end{figure*}

We conducted user studies on the Tencent Questionnaire platform to evaluate the facial reactions generated by our proposed method, ReactDiff, in comparison to four state-of-the-art methods: Zhou et al. \cite{zhou2022responsive}, Ng et al \cite{ng2022learning}, ReactFace \cite{luo2023reactface} and ground truth (GT) real facial reactions.
The designed user interface is shown in Fig.~\ref{fig:use_study}.
Specifically, 21 volunteers (seven females, 14 males) with expertise in machine learning or deep learning participated in an online survey aimed at determining their preferences between facial reaction sequences generated by ReactDiff and the competitor methods. 
Each volunteer watched eight video clips (24 sequence group pairs total), with each clip showing two groups of generated reaction sequences to the same speaker video, one group of reactions generated by ReactDiff, and one by a competitor method. The sequences were randomized and volunteers evaluated the quality of generated reactions on realism, diversity, appropriateness, and synchronization. As shown in Tab.~\ref{ex:user_study}, reactions by our proposed ReactDiff method were preferred by over 69.1\% of participants in most cases when compared to Zhou et al. and Ng et al. ReactDiff also achieved superior results to ReactFace. Interestingly, ReactDiff produced reactions close in quality to the ground truth reactions.

\subsection{Dataset Coverage and Multilingual Performance}
\label{subsec:multilingual}

\noindent\textbf{Coverage.}
The \textsc{REACT2024} corpus already spans diverse interaction contexts and cultures: 133 participants recorded across sites in France, Germany, and the UK; conversations in English, Spanish, Italian, Indonesian, French; and more than 58 topics (\emph{e.g.}, travel, technology, health, cooking, sports, video games). Scenarios include knowledge transfer, information retrieval, and task interruptions.

\medskip\noindent\textbf{Multilingual results.}
We report per-language results on \textsc{REACT2024} (English, French, German). As shown in Tab.~\ref{tb:multilingual}, ReactDiff consistently improves diversity (FRVar), realism (FVD~$\downarrow$), and synchrony (FRSyn~$\downarrow$) over a diffusion baseline across all three languages, indicating that the model learns language-conditioned reaction patterns while preserving generalization.

\begin{table}[t]
\centering
\caption{\textbf{Multilingual performance} on \textsc{REACT2024}.}
\resizebox{0.48\textwidth}{!}{
\begin{tabular}{l|ccc|ccc|ccc}
\toprule
\multirow{2}{*}{Method} &
\multicolumn{3}{c|}{English} &
\multicolumn{3}{c|}{French} &
\multicolumn{3}{c}{German} \\
& FRVar $\uparrow$ & FVD $\downarrow$ & FRSyn $\downarrow$
& FRVar $\uparrow$ & FVD $\downarrow$ & FRSyn $\downarrow$
& FRVar $\uparrow$ & FVD $\downarrow$ & FRSyn $\downarrow$ \\
\midrule
Diffusion model & 0.012 & 394.4 & 48.4 & 0.013 & 413.7 & 47.1 & 0.020 & 385.6 & 46.5 \\
\rowcolor{Gray}
ReactDiff & \textbf{0.020} & \textbf{386.5} & \textbf{42.7} & \textbf{0.020} & \textbf{398.0} & \textbf{41.5} & \textbf{0.026} & \textbf{372.5} & \textbf{43.9} \\
\bottomrule
\end{tabular}
}
\label{tb:multilingual}
\end{table}

\subsection{Runtime, Model Size, and Distillation}
\label{subsec:efficiency}

We measure efficiency on a single NVIDIA GeForce GTX~1080~Ti (11\,GB). With 50 denoising steps, ReactDiff reaches 10.4~FPS. Reducing to 10 steps yields 36.9~FPS with a modest trade-off in FRDiv/FRCorr. After model distillation (83.95M $\rightarrow$ 19.87M params), the 10-step model peaks at 42.3~FPS while remaining competitive on quality metrics (Tab.~\ref{tb:efficiency}).

\begin{table}[t]
\centering
\caption{\textbf{Efficiency vs.\ quality.} Results on a single GTX~1080~Ti (11\,GB).}
\resizebox{0.48\textwidth}{!}{
\begin{tabular}{lcccccc}
\toprule
Method & Steps & Params (M) & GFLOPs & FPS $\uparrow$ & FRDiv $\uparrow$ & FRCorr $\uparrow$ \\
\midrule
ReactDiff & 50 & 83.95 & 669.88 & 10.4 & 0.16 & 0.52 \\
ReactDiff & 10 & 83.95 & 133.98 & 36.9 & 0.11 & 0.45 \\
\rowcolor{Gray}
Distilled ReactDiff & 10 & \textbf{19.87} & \textbf{40.50} & \textbf{42.3} & 0.14 & 0.30 \\
\bottomrule
\end{tabular}
}
\label{tb:efficiency}
\end{table}

\subsection{Details of Facial Action Unit Pairs}
\label{subsec:details_facial_action_unit_pairs}

To enhance the naturalness of facial reactions, we introduced a facial action constraint into the diffusion process, elaborated upon in Section 4.2. This constraint supplements the priors governing relationships between human facial action units. Our study identifies three fundamental types of dependencies among facial actions, drawing insights from prior research on facial action unit detection \cite{luo2022learning, zhang2018weakly, li2019semantic} and psychological studies \cite{ekman1978facial}. These dependencies are categorized as symmetric, co-occurred, and mutually exclusive actions. Within each category, we delineate facial action unit pairs showing such dependencies. Specifically, we have identified 20 symmetric pairs, 30 co-occurred pairs, and 58 mutually exclusive pairs, as detailed in Table \ref{tab:facial_action_unit_pairs}.

\begin{table*}[t]\centering
 \caption{Facial action unit pairs used in facial action constraint.}
 \resizebox{0.78\textwidth}{!}{
\begin{tabular}{l|l|l|l|l|l}
\toprule 
\multicolumn{2}{c}{Symmetric Pair}    & \multicolumn{2}{c}{Co-occurred Pair} & \multicolumn{2}{c}{Mutually Exclusive Pair} \\ \midrule
browDownLeft       & browDownRight       & browOuterUpLeft   & eyeLookUpLeft     & browDownLeft          & browOuterUpLeft      \\ \midrule
browOuterUpLeft    & browOuterUpRight    & browOuterUpRight  & eyeLookUpRight    & browDownRight         & browOuterUpRight     \\ \midrule
cheekSquintLeft    & cheekSquintRight    & eyeLookDownLeft   & browDownLeft      & browDownLeft          & eyeLookUpLeft        \\ \midrule
eyeBlinkLeft       & eyeBlinkRight       & eyeLookDownRight  & browDownRight     & browDownRight         & eyeLookUpRight       \\ \midrule
eyeLookDownLeft    & eyeLookDownRight    & eyeBlinkLeft      & browDownLeft      & browDownLeft          & eyeWideLeft          \\ \midrule
eyeLookInLeft      & eyeLookInRight      & eyeBlinkRight     & browDownRight     & browDownRight         & eyeWideRight         \\ \midrule
eyeLookOutLeft     & eyeLookOutRight     & eyeWideLeft       & browOuterUpLeft   & browInnerUp           & eyeBlinkLeft         \\ \midrule
eyeLookUpLeft      & eyeLookUpRight      & eyeWideRight      & browOuterUpRight  & browInnerUp           & eyeBlinkRight        \\ \midrule
eyeSquintLeft      & eyeSquintRight      & eyeWideLeft       & browInnerUp       & eyeLookDownLeft       & eyeLookUpLeft        \\ \midrule
eyeWideLeft        & eyeWideRight        & eyeWideRight      & browInnerUp       & eyeLookDownRight      & eyeLookUpRight       \\ \midrule
jawLeft            & jawRight            & cheekSquintLeft   & MouthLeft         & eyeLookInLeft         & eyeLookOutLeft       \\ \midrule
mouthDimpleLeft    & mouthDimpleRight    & cheekSquintRight  & MouthRight        & eyeLookInRight        & eyeLookOutRight      \\ \midrule
mouthFrownLeft     & mouthFrownRight     & cheekSquintLeft   & mouthSmileLeft    & eyeLookInLeft         & eyeSquintLeft        \\ \midrule
MouthLeft          & MouthRight          & cheekSquintRight  & mouthSmileRight   & eyeLookInRight        & eyeSquintRight       \\ \midrule
mouthLowerDownLeft & mouthLowerDownRight & cheekSquintLeft   & mouthFrownLeft    & eyeWideLeft           & eyeBlinkLeft         \\ \midrule
mouthPressLeft     & mouthPressRight     & cheekSquintRight  & mouthFrownRight   & eyeWideRight          & eyeBlinkRight        \\ \midrule
mouthSmileLeft     & mouthSmileRight     & cheekSquintLeft   & mouthDimpleLeft   & jawOpen               & mouthClose           \\ \midrule
mouthStretchLeft   & mouthStretchRight   & cheekSquintRight  & mouthDimpleRight  & mouthClose            & mouthUpperUpLeft     \\ \midrule
mouthUpperUpLeft   & mouthUpperUpRight   & cheekSquintLeft   & mouthUpperUpLeft  & mouthClose            & mouthUpperUpRight    \\ \midrule
noseSneerLeft      & noseSneerRight      & cheekSquintRight  & mouthUpperUpRight & mouthClose            & tongueOut            \\ \midrule
                   &                     & cheekSquintLeft   & mouthPressLeft    & mouthClose            & mouthLowerDownLeft   \\ \midrule
                   &                     & cheekSquintRight  & mouthPressRight   & mouthClose            & mouthLowerDownRight  \\ \midrule
                   &                     & browOuterUpLeft   & mouthSmileLeft    & mouthFrownLeft        & mouthSmileLeft       \\ \midrule
                   &                     & browOuterUpRight  & mouthSmileRight   & mouthFrownRight       & mouthSmileRight      \\ \midrule
                   &                     & noseSneerLeft     & cheekSquintLeft   & mouthFrownLeft        & mouthUpperUpLeft     \\ \midrule
                   &                     & noseSneerRight    & cheekSquintRight  & mouthFrownRight       & mouthUpperUpRight    \\ \midrule
                   &                     & mouthFrownLeft    & browDownLeft      & mouthFrownLeft        & mouthFunnel          \\ \midrule
                   &                     & mouthFrownRight   & browDownRight     & mouthFrownRight       & mouthFunnel          \\ \midrule
                   &                     & mouthUpperUpLeft  & browDownLeft      & mouthFrownLeft        & mouthFrownRight      \\ \midrule
                   &                     & mouthUpperUpRight & browDownRight     & mouthFunnel           & mouthLowerDownLeft   \\  \midrule
                   &                     &                   &                   & mouthFunnel           & mouthLowerDownRight  \\  \midrule
                   &                     &                   &                   & mouthFunnel           & mouthRollLower       \\  \midrule
                   &                     &                   &                   & mouthFunnel           & mouthSmileLeft       \\  \midrule
                   &                     &                   &                   & mouthFunnel           & mouthSmileRight      \\  \midrule
                   &                     &                   &                   & mouthFunnel           & tongueOut            \\ \midrule
                   &                     &                   &                   & mouthLowerDownLeft    & mouthPressLeft       \\ \midrule
                   &                     &                   &                   & mouthLowerDownRight   & mouthPressRight      \\ \midrule
                   &                     &                   &                   & mouthLowerDownLeft    & mouthUpperUpLeft     \\ \midrule
                   &                     &                   &                   & mouthLowerDownRight   & mouthUpperUpRight    \\ \midrule
                   &                     &                   &                   & mouthLowerDownLeft    & mouthSmileLeft       \\ \midrule
                   &                     &                   &                   & mouthLowerDownRight   & mouthSmileRight      \\ \midrule
                   &                     &                   &                   & mouthPressLeft        & mouthStretchLeft     \\ \midrule
                   &                     &                   &                   & mouthPressRight       & mouthStretchRight    \\  \midrule
                   &                     &                   &                   & mouthPressLeft        & mouthUpperUpLeft     \\  \midrule
                   &                     &                   &                   & mouthPressRight       & mouthUpperUpRight    \\  \midrule
                   &                     &                   &                   & mouthPucker           & jawOpen              \\  \midrule
                   &                     &                   &                   & mouthFunnel           & jawOpen              \\  \midrule
                   &                     &                   &                   & mouthPucker           & mouthUpperUpLeft     \\  \midrule
                   &                     &                   &                   & mouthPucker           & mouthUpperUpRight    \\  \midrule
                   &                     &                   &                   & mouthPucker           & mouthLowerDownLeft   \\  \midrule
                   &                     &                   &                   & mouthPucker           & mouthLowerDownRight  \\  \midrule
                   &                     &                   &                   & mouthRollLower        & mouthRollUpper       \\  \midrule
                   &                     &                   &                   & mouthShrugLower       & mouthShrugUpper      \\  \midrule
                   &                     &                   &                   & mouthShrugLower       & tongueOut            \\  \midrule
                   &                     &                   &                   & mouthSmileLeft        & mouthStretchLeft     \\  \midrule
                   &                     &                   &                   & mouthSmileRight       & mouthStretchRight    \\  \midrule
                   &                     &                   &                   & mouthSmileLeft        & mouthUpperUpLeft     \\  \midrule
                   &                     &                   &                   & mouthSmileRight       & mouthUpperUpRight   \\
\bottomrule
\end{tabular}}

\label{tab:facial_action_unit_pairs}
\end{table*}

\end{document}